\documentclass{article} %
\usepackage{iclr2025_conference,times}

\usepackage{amsmath,amsfonts,bm}

\def\eqref#1{equation~\ref{#1}}

\def\1{\bm{1}}

\DeclareMathAlphabet{\mathsfit}{\encodingdefault}{\sfdefault}{m}{sl}
\SetMathAlphabet{\mathsfit}{bold}{\encodingdefault}{\sfdefault}{bx}{n}

\DeclareMathOperator*{\argmin}{arg\,min}

\usepackage{url}

\usepackage{algorithm}  
\usepackage{algorithmic}  
\usepackage{amsmath}  
\usepackage{amsfonts}  
\usepackage{amssymb} 
\usepackage{macro_commands}
\usepackage{tcolorbox}
\tcbuselibrary{skins}
\usepackage{wrapfig}
\usepackage{tablefootnote}
\definecolor{ccr}{RGB}{10,110,150}  
\usepackage{enumitem}

\definecolor{linkColor}{rgb}{0.2,0.4,0.6}
\usepackage[utf8]{inputenc} 
\usepackage[T1]{fontenc}   
\usepackage[colorlinks=true,linkcolor=linkColor,citecolor=linkColor,filecolor=linkColor,urlcolor=linkColor]{hyperref} 

\newtcolorbox{prompt}[1]{
    enhanced,
    drop shadow=black!5!white,
    left=4mm,
    right=4mm,
    top=1mm,
    bottom=1mm,
    boxsep=0mm,
    rounded corners,
    title=#1,
    fontupper=\scriptsize\linespread{0.7}\fontfamily{lmr}\selectfont,
    }

\newcommand{\methodName}{\textsc{SynPO}}

\allowdisplaybreaks
\usepackage{colortbl}
\definecolor{lightgray}{gray}{0.9} %
\definecolor{lightgray1}{gray}{0.9} %
\definecolor{lightgray2}{gray}{0.84} %
\definecolor{lightgray3}{gray}{0.77} %
\definecolor{lightgray4}{gray}{0.7} %

\def \algname {\text{Synthetic Preference Optimization}}

\title{Self-Boosting LLMs with Synthetic Preference Data}
\title{Self-Boosting Large Language Models with \\ Synthetic Preference Data}

\author{
Qingxiu Dong$^{\dagger,\ddagger}$~~~\ Li Dong$^\ddagger$~~~\ Xingxing Zhang$^\ddagger$~~~\ Zhifang Sui$^\dagger$~~~\ Furu Wei$^\ddagger$ \\ 
$^\dagger$Peking University~~~~\ $^\ddagger$Microsoft Research 
\\
}

\iclrfinalcopy %
\begin{document}

\maketitle

\begin{abstract}
Through alignment with human preferences, Large Language Models (LLMs) have advanced significantly in generating honest, harmless, and helpful responses.
However, collecting high-quality preference data is a resource-intensive and creativity-demanding process, especially for the continual improvement of LLMs.
We introduce SynPO, a self-boosting paradigm that leverages synthetic preference data for model alignment.
SynPO employs an iterative mechanism wherein a self-prompt generator creates diverse prompts, and a response improver refines model responses progressively.
This approach trains LLMs to autonomously learn the generative rewards for their own outputs and  eliminates the need for large-scale annotation of prompts and human preferences.
After four SynPO iterations, Llama3-8B and Mistral-7B show significant enhancements in instruction-following abilities, achieving over 22.1\% win rate improvements on AlpacaEval 2.0 and ArenaHard.
Simultaneously, SynPO improves the general performance of LLMs on various tasks, validated by a 3.2 to 5.0 average score increase on the well-recognized Open LLM leaderboard.

\end{abstract}

\section{Introduction}

Large Language Models (LLMs) have made remarkable progress in following user instructions and generating honest, harmless, and helpful responses~\citep{achiam2023gpt,Llama31}. This advancement is primarily achieved in the model alignment stage, which involves training reward models or LLMs directly on datasets curated from human preferences~\citep{Ouyang2022TrainingLM,bai2022training}
, typically employing Reinforcement Learning from Human Feedback (RLHF)~\citep{Ouyang2022TrainingLM} or Direct Preference Optimization (DPO)~\citep{dpo}.

Recent research has made significant strides in model alignment by 
collecting high-quality preference data~\citep{Hu2024TowardsCP}
, sampling and ranking on-policy responses~\citep{meng2024simpo,sppo}, or introducing LLM-as-a-Judge as substitutes for human preferences~\citep{Yuan2024SelfRewardingLM,cui2023ultrafeedback}.  
However, most work still relies on static, pre-collected preference datasets from human or stronger LLM annotation.
As LLMs improve rapidly, collecting large, high-quality preference data for effective learning becomes increasingly challenging and costly, whether from humans or stronger models~\citep{Shi2023SaferInstructAL}.
According to \citet{sapo}, directly sampling preference pairs, which closely resembles an on-policy setting, can result in performance declines due to inherent volatility and inefficiency.
Therefore, constructing effective preference data to continuously improve LLMs remains a critical research problem.

In this work, we present a self-boosting paradigm for LLM alignment, SynPO.
This paradigm leverages a small set of supervised fine-tuning (SFT) data to steer the generation of synthetic preference data, thereby enabling LLMs to iteratively extend their capabilities through optimizing on synthetic data.  
To support iterative preference learning across diverse scenarios, SynPO first trains a self-prompt generator to create large-scale synthetic prompts. 
Unlike previous approaches that require more powerful LLMs and instruction examples~\citep{Wang2022SelfInstructAL}, our generator utilizes only the LLM itself and three random keywords as input.
To generate preference pairs for the synthetic prompts,
SynPO utilizes model-generated responses as rejected candidates and employs a response improver to refine these responses into chosen ones.
The response improver comes from two straightforward intuitions: (1) LLMs excel at identifying distribution gaps between texts \citep{zhong2022describing, singh2022explaining}, and (2) refining a response is generally easier than generating a high-quality response from scratch~\citep{Madaan2023SelfRefineIR,jqself,Nguyen2024BetterAW}.
In each iteration, we train the initial model to be a response improver, focusing on identifying distribution gaps between current model outputs and gold standard responses in seed data. We then use the response improver to refine the model outputs, thereby providing generative rewards to the responses. 
This approach allows the LLM to make subtle improvements and gradually push its boundaries.
By leveraging small high-quality data and the current model state to guide the generation of synthetic data, we introduce stronger supervision in an iterative manner. 
  
\begin{figure}[t]  
    \centering  
    \begin{minipage}[b]{0.475\textwidth}  
        \centering  
        \includegraphics[width=\linewidth]{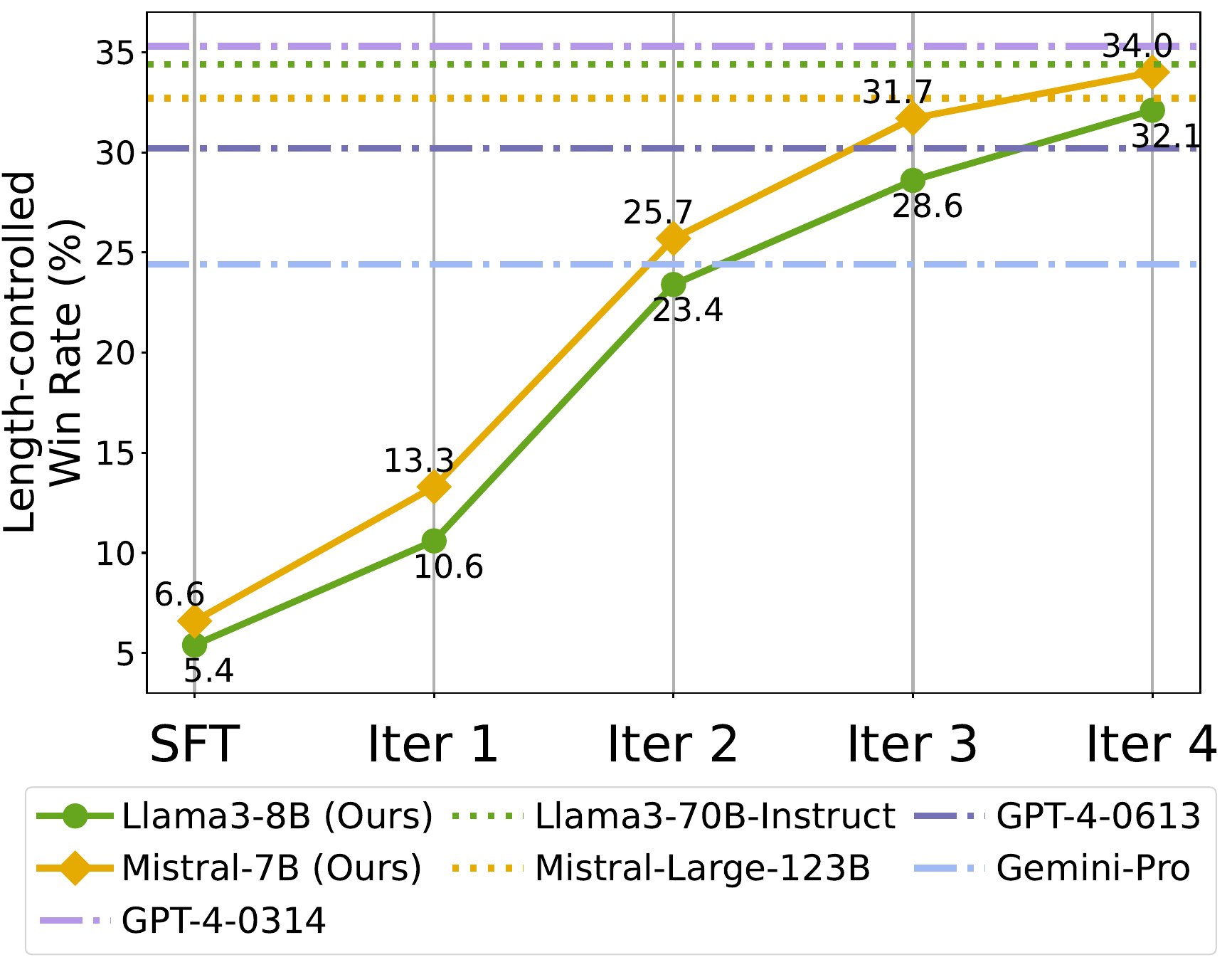}  
        \vspace{-15pt}
        \caption{Length-controlled win rate on AlpacaEval 2.0 improves with SynPO iterations, approaching GPT-4 level for the base versions of Llama3-8B and Mistral-7B.} 
        \label{fig:improving}  
    \end{minipage}%
    \hfill  
    \begin{minipage}[b]{0.48\textwidth}  
        \centering  
        \small  
        \setlength{\tabcolsep}{2pt}  
        \resizebox{0.99\textwidth}{!}{ %
\begin{tabular}{l|rrr}                    
    \toprule           
    \bf Model & \bf Size & \bf LC (\%) & \bf WR (\%) \\             
    \midrule            
    gpt4\_1106\_preview & - & 50.0 & 50.0 \\            
    GPT-4 (03/14) & - & 35.3 & 22.1 \\            
    Meta-Llama-3-70B-Instruct & 70B & 34.4 & 33.2 \\            
    \rowcolor{lightgray4} \bf Mistral-Base-SynPO \hspace{0.5em}\textit{Iter4} & \bf 7B & \bf 34.0 & \bf 36.4 \\            
    Mistral Large (24/02) & 123B & 32.7 & 21.4 \\            
    \rowcolor{lightgray3} \bf Mistral-Base-SynPO \hspace{0.5em}\textit{Iter3} & \bf 7B & \bf 31.7 & \bf 33.8 \\            
    GPT-4 (06/13) & - & 30.2 & 15.8 \\            
    Claude 2 & - & 28.2 & 17.2 \\            
    Claude 2.1 & - & 27.3 & 17.0 \\            
    \rowcolor{lightgray2} \bf Mistral-Base-SynPO \hspace{0.5em}\textit{Iter2} & \bf 7B & \bf 25.7 & \bf 28.1 \\            
    gemini-pro & - & 24.4 & 18.2 \\            
    Mixtral-8x7B-Instruct-v0.1 & 8x7B & 23.7 & 18.3 \\            
    Mistral-7B-Instruct-v0.2 & 7B & 17.1 & 14.7 \\            
    \rowcolor{lightgray1} \bf Mistral-Base-SynPO \hspace{0.5em}\textit{Iter1} & \bf 7B & \bf 13.3 & \bf 15.3 \\            
    Mistral-Base-SFT & 7B & 6.6 & 3.6 \\            
    \bottomrule        
\end{tabular}  

        }  
        \makeatletter\def\@captype{table}\makeatother\caption{Results on AlpacaEval 2.0 leaderboard. LC and WR represent length-controlled and raw win rate, respectively.}
        \label{tab:improving}  
    \end{minipage}  
\end{figure}

Experimental results demonstrate that SynPO not only benefits LLM alignment with human preferences, but also improves generalist capabilities across various tasks.
Trained solely on synthetic data, SynPO significantly improves the instruction-following abilities of Llama3-8B and Mistral-7B (as shown in Figure~\ref{fig:improving} and Table~\ref{tab:improving}), achieving over a 26\% length-controlled win rate improvement on AlpacaEval 2.0~\citep{alpacaeval} and a 22\% to 30\% improvement on Arena-hard~\citep{arenahard} (as shown in Table~\ref{tab:main_res}). Furthermore, self-boosted models achieve 3.2\% to 5.0\% higher average performance than SFT models on the Open LLM leaderboard~\citep{open-llm-leaderboard}, indicating SynPO also enhances general LLM performance. 

To summarize, our contribution includes:
\begin{itemize}[leftmargin=*]
\setlength\itemsep{0.01em}
\item We introduce SynPO, a self-boosting mechanism that iteratively induces LLMs to synthesize high-quality data for training.  
Without requiring human-annotated preference data, SynPO significantly enhances the diversity and quality of synthetic prompts and responses. 
\item  SynPO dynamically guides LLMs to improve their own outputs, using pre- and post-refinement generations as synthetic preference pairs for training. This approach effectively integrates generative rewards for preference learning, enabling LLMs to gradually push their boundaries.
\item  SynPO significantly enhances both the instruction-following capabilities and the general performance of LLMs, showing substantial improvements over three to four iterations.
\end{itemize}

\section{Self-Boosting LLM with Synthetic Preference Data}

\begin{figure}
    \centering
    \includegraphics[width=0.95\linewidth]{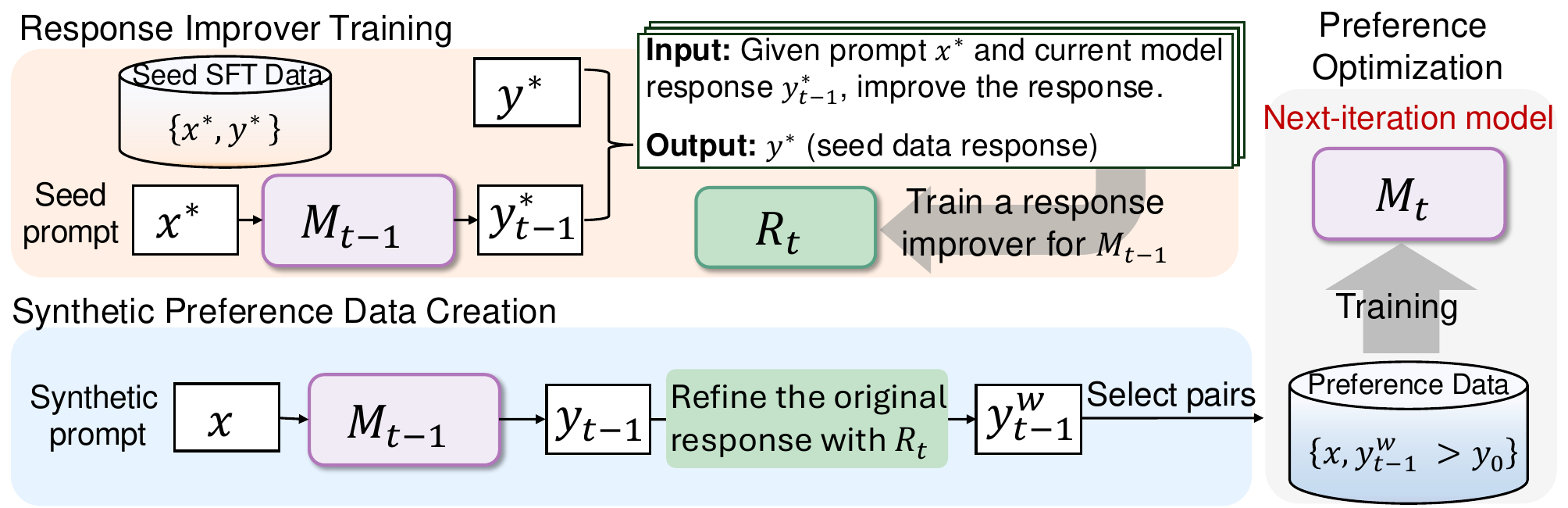}
    \caption{\textbf{Overview of SynPO in the $\mathbf{t^{th}}$ iteration.} Starting with the previous iteration model $M_{t-1}$, SynPO first learns a response improver $R_t$ to identify discrepancies between model responses ($y^*_{t-1}$) and gold standard responses ($y^*$) on seed data, and learns to refine model responses. Subsequently, on the self-generated prompts $x$ (elaborated in Section~\ref{method:prompt-gen}), SynPO employs $R_t$ to refine the $M_{t-1}$ responses ($y_{t-1}$) into improved responses ($y_{t-1}^w$). The valid synthetic prompts $x$, refined responses ($y_{t-1}^w$), and initial model $M_{0}$ responses ($y_0$) to form synthetic preference data. These data are incorporated into the synthetic preference dataset for preference optimization, resulting in an updated $M_t$ for the next iteration. The iterative process continually enhances LLM capabilities in instruction-following and task performance.  
}
    \label{fig:main}
\end{figure}

SynPO is a self-boosting scheme designed to iteratively generate high-quality preference data. An overview of SynPO is presented in Figure \ref{fig:main}.
It begins with a small set of SFT data as seed data, denoted as $\left\{(\mathbf{x}^*_i, \mathbf{y}^*_i)\right\}_{i=0}^n$, and the initial policy model $\pi_{\theta_0}$.
By incorporating both the self-prompt generator and the response improver, SynPO provides sufficient prompts for iterative training and leverages the generative rewards in the synthetic preference data. 
This approach allows the policy model to make subtle improvements and gradually expand its boundaries.

\subsection{Synthetic Prompt Creation}
\label{method:prompt-gen}
Diverse and ample prompts are crucial for effective preference learning~\citep{Shi2023SaferInstructAL,Yuan2024SelfRewardingLM,Song2024ScalingDD}. Diversity facilitates generalization and a sufficient number of prompts allows data selection from a large candidate pool. 
In SynPO, we propose a novel strategy for synthetic prompt generation.
We design a keywords-to-text task to guide the training of a self-prompt generator and create pseudo-label data from the seed SFT data.

\begin{figure}[t]
    \centering
    \input{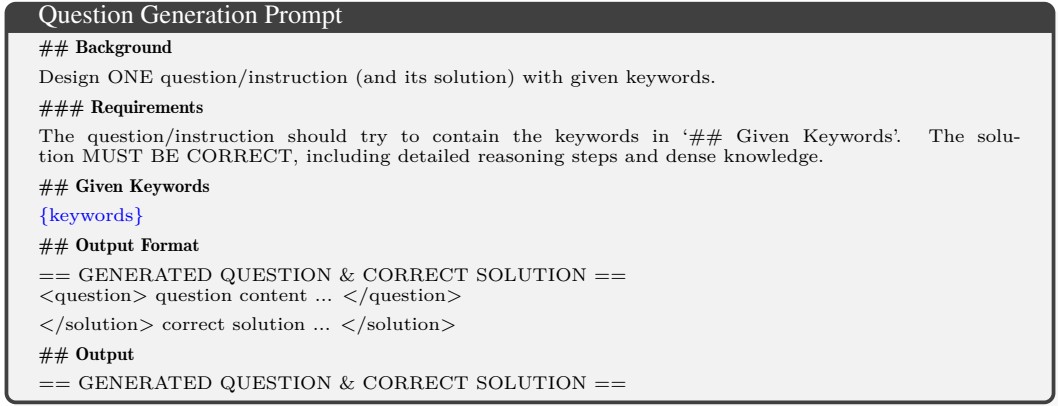}
    \caption{Prompt used in SynPO for LLMs to act as self-prompt generators.
    }
    \label{fig:qgen_prompt}
\end{figure}

\paragraph{Self-Prompt Generator Training}
We train the LLM itself to serve as a high-quality prompt generator. 
For each prompt $\mathbf{x}^*_i$ in seed data, we randomly extract two keywords from $\mathbf{x}^*_i$ and one noise keyword from $\mathbf{x}^*_j$, where $j \in \{1, 2, \ldots, n\} \setminus \{i\}$.
The inclusion of the noise keyword enhances the robustness of the prompt generator. It learns to filter out irrelevant keywords during training and ensure that the generated prompts are fluent.
This process yields a keyword list, $k_i$ for $\mathbf{x}^*_i$.
Next, we insert $k_i$ into a prompt template (see Figure~\ref{fig:qgen_prompt}) to create a prompt and use $(\mathbf{x}^*_i, \mathbf{y}^*_i)$ as the corresponding completion. This process constructs training data for the prompt generator.  
We then optimize $\theta_0$ through SFT to transform the model into a prompt generator $\mathcal{G}$. 
$\mathcal{G}$ possesses the capability to generate unlimited, diverse, and high-quality user instructions, controlled by the given keywords.

\paragraph{Keywords Sampling and Prompt Generating}
\label{sec:similarity-distribution}
To enhance the overall diversity of prompts, we sample keywords from pretrain corpus paragraphs.
We select three keywords from a same paragraph to maintain the inherent distribution between keywords, and sample from different paragraphs to ensure overall diversity.
Specifically, we randomly sample keyword lists from the RefinedWeb paragraphs~\citep{refinedweb} and generate $m$ synthetic prompts, denoted $\{\mathbf{x}_i\}_{i=1}^{m}$, $\mathbf{x}_i \sim \mathcal{G}(\cdot|\mathbf{k}_i)$. 

Benefiting from the self-prompt generator, our approach requires only a small SFT data to enable the model to generate diverse prompts independently, no in-context learning examples or predefined topic lists are required.
Due to the high combinatorial possibilities of keyword sampling, the self-prompt generator can produce an infinite variety of synthetic prompts for preference learning.

\begin{figure}[t]  
    \centering  
    \begin{minipage}[b]{0.47\textwidth}  
        \centering  
        \includegraphics[width=\linewidth]{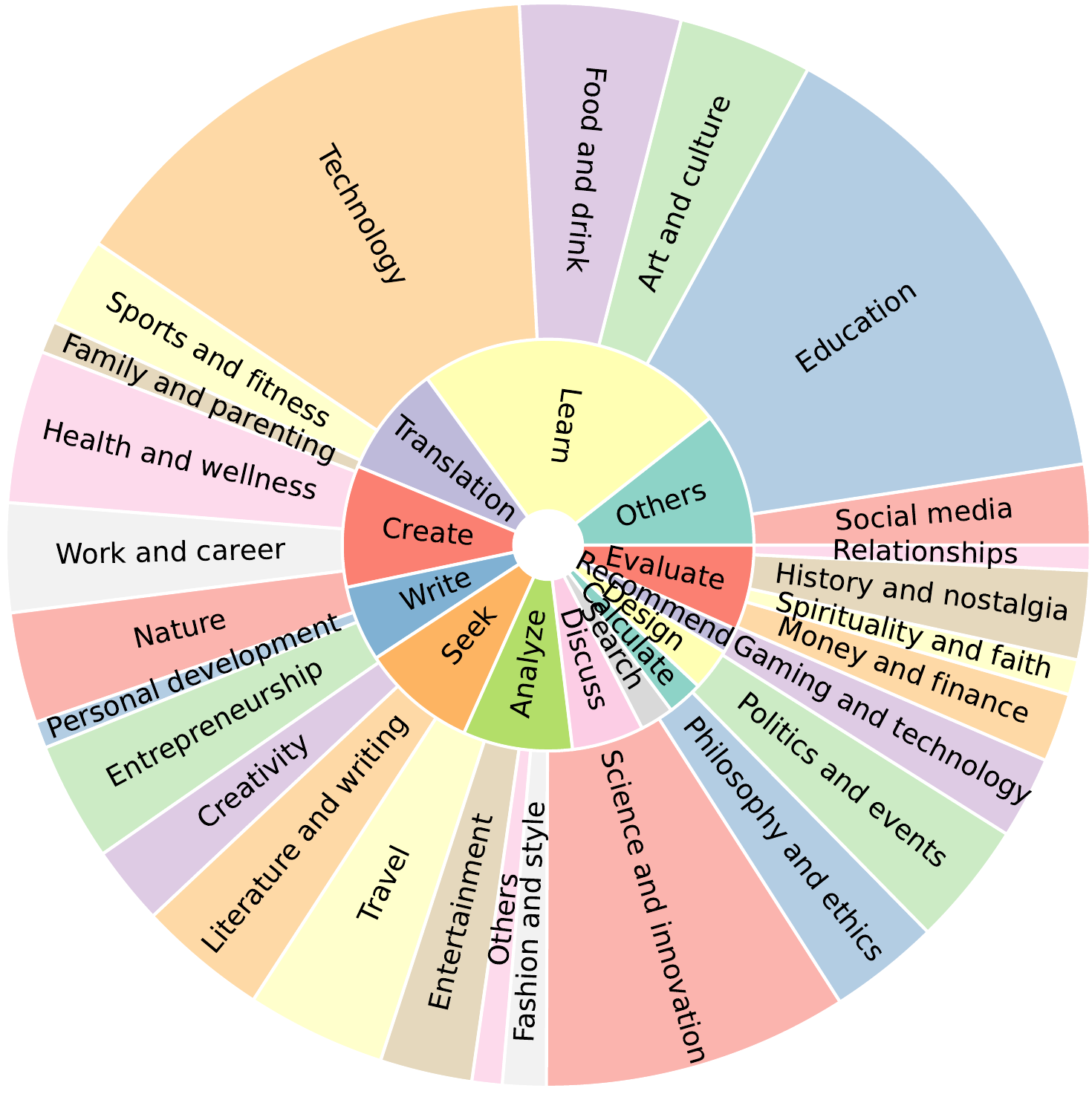}  
        \caption{The top 25 most common topics (outer circle) and the top 12 most common intentions (inner circle) in SynPO generated prompts. We aggregate the other topics and intentions to the `Others' group. 
}  
        \label{fig:diverse-topics}  
    \end{minipage}\hfill  
    \begin{minipage}[b]{0.48\textwidth}  
        \centering  
        \includegraphics[width=\linewidth]{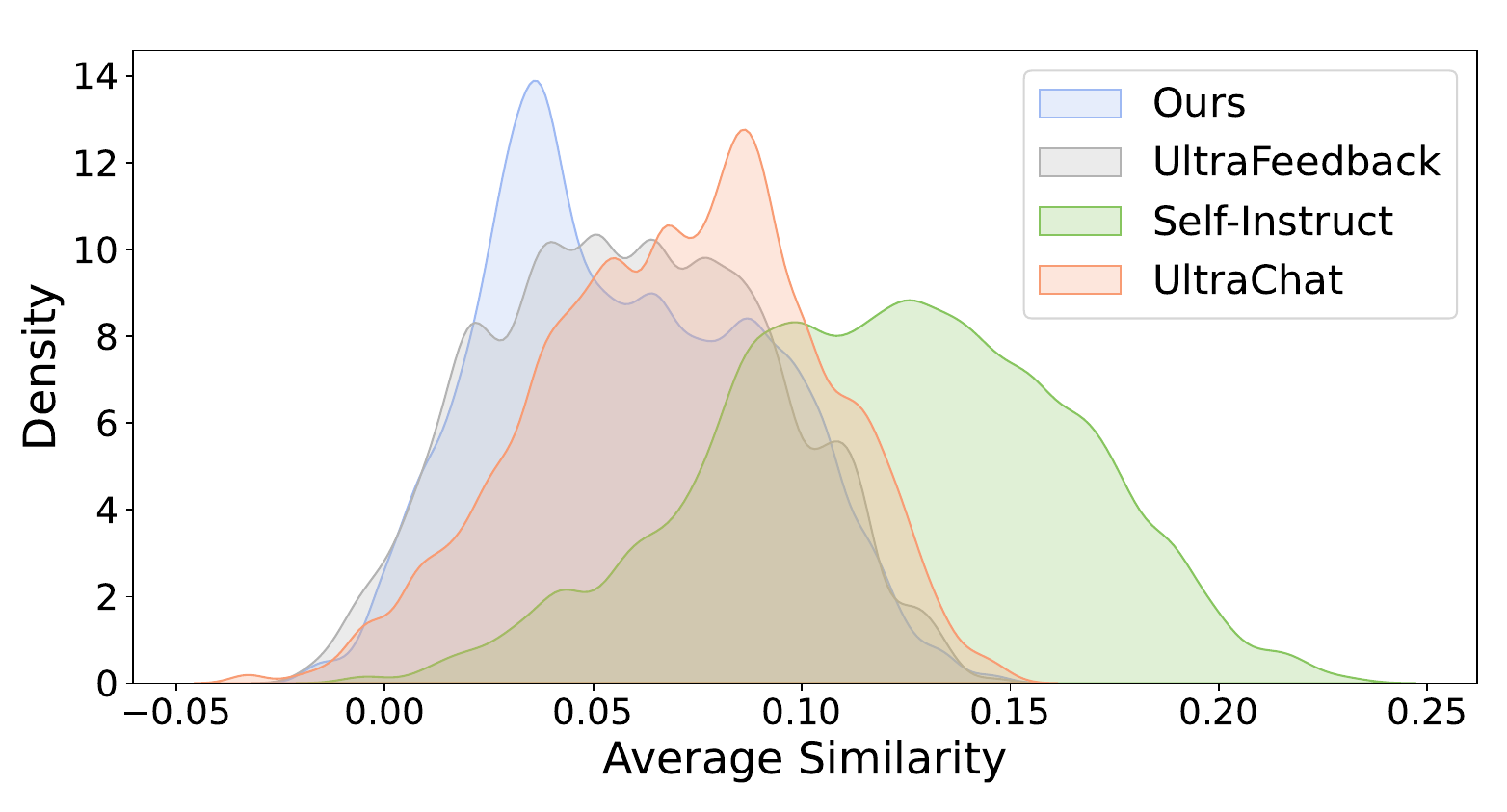}  
        \caption{Inter-prompt similarity distributions for 1,000 randomly sampled prompts from SynPO, UltraFeedback~\citep{cui2023ultrafeedback}, Self-Instruct~\citep{Wang2022SelfInstructAL}, and UltraChat~\citep{ding2023enhancing}. We used SentenceTransformer ~\citep{reimers-2019-sentence-bert} to compute sentence embeddings and calculated the cosine similarity between each prompt and all others, then averaged these values for each prompt. The results suggest that our method, SynPO,  generates more diverse prompts with lower inter-prompt similarity.}
        \label{fig:similarity-distribution}  
    \end{minipage}  
\end{figure}

For further analysis, we randomly sampled 1k self-generated prompts from the Llama3-8B-Base prompt generator and used GPT-4 Turbo to classify the intentions and topics behind these prompts\footnote{Experimental details in Appendix~\ref{app:prompt-analysis}.}.
The results, shown in Figure~\ref{fig:diverse-topics}, demonstrate significant diversity across various topics and user intentions.  
Even compared to prompts from GPT3.5-Turbo~\citep{ding2023enhancing} or a collection of prompts from different sources~\citep{cui2023ultrafeedback}, as shown in Figure~\ref{fig:similarity-distribution}, SynPO generated prompts exhibit lower inter-prompt similarity and greater diversity.  

\subsection{Synthetic Preference Generation}
A primary challenge in leveraging synthetic prompts is the lack of high-quality responses to provide sufficient supervision~\citep{Li2024SyntheticD}. 
To address this, we introduce a response improver to enhance the quality of model responses to synthetic prompts.
Pre- and post-improvement responses naturally become the rejected and chosen candidates, respectively, with the chosen ones providing clear guidance on what approximates a gold standard response.

\paragraph{Response Improver Training}
In each iteration, we train the LLM as a response improver to further reduce the gap between policy model outputs and gold standard responses.   
Formally, let $\pi_{\btheta_{t-1}}$ denote the policy model at the beginning of the $t$-th iteration. We generate outputs from $\pi_{\btheta_{t-1}}$ for the seed data prompts: $\mathbf{y}^*_{(t-1),i} \sim \pi_{\btheta_{t-1}}(\cdot|\mathbf{x}^*_i)$, $i \in \{1, \dots, m\}$. These outputs, along with the seed data responses, form the training set for the response improver, following the template provided in Appendix~\ref{app:rwt_prompt}. Each training example consists of the prompt and the policy model output $(\mathbf{x}^*_i, \mathbf{y}^*_{(t-1),i})$ as the input, and the gold standard response $\mathbf{y}^*_i$ as the output.  
We fine-tune $\pi_{\btheta_0}$ on the training set to obtain the response improver $\mathcal{R}_t$. This response improver refines the policy model outputs, aligning them more closely with the gold standard responses.

\paragraph{Response Improving}
Subsequently, we use $\mathcal{R}_t$ to refine model responses to synthetic prompts, obtaining pre- and post-improvement responses as synthetic preference pairs. For each synthetic prompt $\mathbf{x}_i$, we first obtain the current model output $\mathbf{y}_{(t-1),i} \sim \pi_{\btheta_{t-1}}(\cdot|\mathbf{x}_i)$, for $i \in \{1, \dots, m\}$. The response improver then refines this completion to produce $\overline{\mathbf{y}_{(t-1),i}} \sim \mathcal{R}_t(\cdot|\mathbf{x}_i, \mathbf{y}_{(t-1),i})$, considered the chosen response.
As we fine-tune the initial model in each iteration, the initial policy model output $\mathbf{y}_{(0),i}$ serves as the on-policy rejected response for $\mathbf{x}_i$. Here, $\mathbf{y}_{(0),i} \sim \pi_{\btheta_0}(\cdot|\mathbf{x}_i)$.   
This method generates numerous synthetic preference candidates, including both chosen and rejected responses.

\paragraph{Data Filtering}
Unlike data from humans or strong teacher LLMs, which come with clear standard responses, self-generated data require proper filtering to ensure quality~\citep{rest}. 
As our policy model improves, many responses no longer need refining; we only need to retain data with a preference gap between the chosen and rejected responses.   
Instead of using GPT4-Turbo-as-a-Judge for data filtering~\citep{Rosset2024DirectNO}, SynPO employs only a small model (e.g., a 0.4B PairRM~\citep{pairrm}) or the model itself for scoring. Similar to SimPO~\citep{meng2024simpo} and SPPO~\citep{sppo}, this ensures the process does not rely on a more powerful teacher model. 
We retain $\overline{\mathbf{y}_{(t-1),i}}$ and $\mathbf{y}_{(0),i}$ with significant preference differences (i.e., a large score gap). $\overline{\mathbf{y}_{(t-1),i}}$ is regarded as the chosen response, $\mathbf{y}_{i}^w$, while $\mathbf{y}_{(0),i}$ is regarded as the rejected response, $\mathbf{y}_{i}^l$. Along with the corresponding prompt $\mathbf{x}_i$, they form a valid instance $(\mathbf{x}_i, \mathbf{y}_{i}^w, \mathbf{y}_{i}^l)$.  
All valid data is then integrated into the synthetic preference data for subsequent iterations.

\subsection{Synthetic Preference Optimization}
The large-scale synthetic preference data naturally facilitate the multi-iteration process of self-boosting. 
In each iteration, we follow SimPO~\citep{meng2024simpo} for training; actually, our method is also compatible with other preference optimization training methods, such as DPO~\citep{dpo} and KTO~\citep{kto}. Denoting $\mathcal{D}$ as the synthetic preference data, we have:
\begin{align*}  
    \btheta_{t} \leftarrow \argmin_{\btheta} \mathbb{E}_{(\mathbf{x}_i, \mathbf{y}_{i}^w, \mathbf{y}_{i}^l) \sim \mathcal{D}} \left[  
        \log \sigma \left(  
            \frac{\beta}{|\mathbf{y}_{i}^w|} \log \pi_{\theta_{t-1}} (\mathbf{y}_{i}^w \mid \mathbf{x}_i)  
            - \frac{\beta}{|\mathbf{y}_{i}^l|} \log \pi_{\theta_{t-1}} (\mathbf{y}_{i}^l \mid \mathbf{x}_i)  
            - \gamma  
        \right)  
    \right]  
\end{align*} 
$\sigma$ and $\gamma$ are hyperparameters.
Different from the vanilla SimPO, SynPO is a iterative process and all the preference data are synthetic ones. 
The response improver continuously refines the generation distribution to align with the ideal data distribution across multiple iterations.

Overall, the response improver automatically learns to generate implicit generative rewards for the outputs of the LLM.
Unlike using a discriminative reward model straightforwardly, this approach helps the model learning to improve its outputs.
We present the \textit{\algname} algorithm in Appendix~\ref{app:synpo}. 
The entire optimization process is performed on synthetic preference data, requiring only a small amount of high-quality data for validation. This strategy maintains two key advantages: (1) Compared to the limited and hard-to-collect preference data, SynPO generates an unlimited amount of new self-synthetic data to meet the needs of iterative model improvement. (2) Using small, high-quality validation data prevents the model from deviating during training and consistently guides the generation of more relevant synthetic data.

\setlength{\tabcolsep}{5pt}
\begin{table*}[!t]
\centering
\resizebox{0.95\textwidth}{!}{
\begin{tabular}{lcccccc}
\toprule
\multirow{3}{*}{\textbf{Data Construction}} & \multicolumn{3}{c}{\textbf{Mistral-Base (7B)}} & \multicolumn{3}{c}{\textbf{Llama3-Base (8B)}} \\ 
\cmidrule(lr){2-4}\cmidrule(lr){5-7}
& \multicolumn{2}{c}{\textbf{AlpacaEval 2.0}} & \multicolumn{1}{c}{\textbf{Arena-Hard}} & \multicolumn{2}{c}{\textbf{AlpacaEval 2.0}} & \multicolumn{1}{c}{\textbf{Arena-Hard}}\\ 
\cmidrule(lr){2-3}\cmidrule(lr){4-4}\cmidrule(lr){5-6}\cmidrule(lr){7-7}
& \bf LC (\%) & \bf WR (\%) &  \bf WR (\%) & \bf LC (\%) & \bf WR (\%) &  \bf WR (\%)  \\
\midrule
SFT & 6.6       & 3.6  &  2.0 & 5.4 & 3.1 &  2.7\\ 
\midrule
Manual Collection &    21.5  &  20.8  &  16.8 &  22.0 & 19.8  &  23.2 \\ 
\midrule
Sampling-Ranking \hspace{0.5em}\textit{Iter1} & 6.5  & 4.4  & 4.1 & 7.2  & 4.3  & 4.4\\
Sampling-Ranking \hspace{0.5em}\textit{Iter2} & 9.3  & 6.4  & 4.3 & 7.7  &  4.7 & 6.2\\
Sampling-Ranking \hspace{0.5em}\textit{Iter3} & 10.6  &  7.5 & 7.9 & 13.8  & 8.2 & 8.4\\
Sampling-Ranking \hspace{0.5em}\textit{Iter4}  & 11.6  & 8.0  & 9.6 & 14.2  &  8.4 & 10.4\\
\midrule
Self-Rewarding \hspace{0.5em}\textit{Iter1} &  19.5 & 19.8 & 11.9 & 20.1  & 20.3 & 20.8\\
Self-Rewarding \hspace{0.5em}\textit{Iter2} &  22.4 & 23.5 & 19.2 & 21.7 & 22.4 & 20.5 \\
Self-Rewarding \hspace{0.5em}\textit{Iter3} & 24.6 & 26.3 & 20.8 & 22.4 & 24.1 & 23.8 \\
Self-Rewarding \hspace{0.5em}\textit{Iter4}  & 26.1 & 28.0 & 21.1 & 24.8 & 25.6 & 25.0 \\
\midrule
\rowcolor{lightgray1} \methodName{} \hspace{0.5em}\textit{Iter1}  & 13.3      & 15.3    &  9.8 &  10.6 & 10.7 &   11.6   \\  
\rowcolor{lightgray2} \methodName{} \hspace{0.5em}\textit{Iter2}  & 25.7      & 28.1   & 20.8  &   23.4  & 24.1  & 24.6   \\  
\rowcolor{lightgray3} \methodName{} \hspace{0.5em}\textit{Iter3}  & 31.7      & 33.8    &  \bf 24.1  &  28.6 & 31.5  & \bf32.5\\ 
\rowcolor{lightgray4} \methodName{} \hspace{0.5em}\textit{Iter4}  & \bf 34.0      & \bf 36.4  &  22.8 &  \bf 32.1 & \bf 33.6  & 31.4   \\ 

\bottomrule
\end{tabular}
}
\caption{Results on AlpacaEval 2.0 and Arena-Hard. LC and WR denote length-controlled and raw win rates, respectively. After four \methodName{} iterations, Mistral-Base and Llama3-Base increase LC by 27.4\% and 26.7\%, respectively, on AlpacaEval 2.0. In Arena-Hard, \methodName{} achieves the highest WR by the third iteration, improving both models by over 22.1\%.  
}
\label{tab:main_res}
\end{table*}

\section{Experiments}

We carry out comprehensive experiments to demonstrate the effectiveness of SynPO in enhancing model alignment and improving general model performance.
\subsection{Experimental Setup}
\paragraph{Models and Training}
We perform synthetic preference optimization on both Mistral-Base 7B and Llama3-8B Base. 
Following \citet{meng2024simpo}, we employ supervised fine-tuned models as the initial models. Specifically, the Mistral-Base 7B model (mistralai/Mistral-7B-v0.1) and the Llama3-8B Base model (meta-llama/Meta-Llama-3-8B-Base) were fine-tuned on the UltraChat-200k dataset as part of the Zephyr~\citep{tunstall2023zephyr} training pipeline.\footnote{\url{https://huggingface.co/alignment-handbook/zephyr-7b-sft-full} and \url{https://huggingface.co/princeton-nlp/Llama-3-Base-8B-SFT}}
Subsequently, we utilize 18k seed data to SFT the self-prompt generator and then generate 50k synthetic prompts per iteration. For data filtering, we employ the 0.4B PairRM~\citep{pairrm} as a small pairwise scoring model for the Mistral-Base 7B. For Llama3-Base 8B, we use Llama3 itself (ArmoRM-Llama3-8B-v0.1\footnote{\url{https://huggingface.co/RLHFlow/ArmoRM-Llama3-8B-v0.1}}) as a scoring model for data filtering, given its superior alignment with human scoring~\citep{meng2024simpo}. 
In each iteration \(t = 1, \ldots, T\), we use model \(\pi_{\theta_{t-1}}\) from the previous iteration to generate synthetic preference data and then preference optimize the initial models again.  
More details on training parameters, filtering thresholds, and implementation environments are provided in Appendix~\ref{app:exp}.

\paragraph{Seed Data Construction}
We randomly sample UltraFeedback~\citep{cui2023ultrafeedback}  prompts and their GPT-4 Turbo completions as our seed data.
The seed data is multipurposely transformed for the training of self-prompt generator, response improver, and the validation of synthetic preference optimization.  
The complete UltraFeedback dataset contains 61k instructions from sources including TruthfulQA~\citep{lin2021truthfulqa}, FalseQA~\citep{Hu2023WontGF}, Evol-Instruct~\citep{xu2023wizardlm}, UltraChat~\citep{ding2023enhancing}, and ShareGPT~\citep{vicuna2023}. 
To construct seed SFT data with high-quality responses, we randomly sampled 18k prompts and obtained the corresponding completions generated by GPT-4 Turbo.  

\paragraph{Baselines}
As baselines, we use the initial supervised fine-tuned models and those optimized with data from various preference construction methods, including manual collection and iterative approaches. 
We recognize the UltraFeedback preference data~\citep{cui2023ultrafeedback} as a product of manual collection. It is gathered from six high-quality datasets and various models, with preferences annotated by GPT-4~\citep{achiam2023gpt}.
For iterative construction approaches, we involve Sampling-Ranking and Self-Rewarding. For Sampling-Ranking, similar to~\citet{meng2024simpo} and~\citet{sppo}, we involves LLMs in sampling five responses per prompt in each iteration. The same scoring models, i.e., PairRM and ArmoRM-Llama3-8B-v0.1, are then used to select the highest and lowest scoring responses as the chosen and rejected responses, respectively.   
For the Self-Rewarding method~\citep{Yuan2024SelfRewardingLM}, we generate preference data based on model's own rewards via LLM-as-a-Judge prompting. We employ our 18k seed data as the initial instruction-following data. Given that Self-Rewarding requires additional LLM-as-a-Judge training data, we generate 16k seed data with GPT-4 Turbo.  
For all settings, we adopt SimPO~\citep{meng2024simpo} for preference optimization. 
More detailed are elaborated in Appendix~\ref{app:bsln}.

\begin{table}[t]  
    \centering  
    \begin{minipage}[b]{0.5\linewidth}  
        \centering  
        \setlength{\tabcolsep}{2pt} 
        \resizebox{0.99\textwidth}{!}{
        \begin{tabular}{lcccc}  
            \toprule  
            \multirow{2}{*}{\bf Data Construction} & \multicolumn{2}{c}{\textbf{Mistral-Base}} & \multicolumn{2}{c}{\textbf{Llama3-Base}} \\  
            \cmidrule(lr){2-3} \cmidrule(lr){4-5}  
            & \bf Turn 1 & \bf Turn 2 & \bf Turn 1 & \bf Turn 2 \\  
            \midrule  
            SFT & 6.04 & 5.65 & 6.55 & 5.36 \\ 
            \midrule
            Manual Collection  & 6.73   & \bf 6.82   & 7.29 & 7.00       \\
            Sampling-Ranking \hspace{0.5em}\textit{Iters*} & 6.83 & 6.18 & 7.06 & 6.99 \\
            Self-Rewarding \hspace{0.5em}\textit{Iters*} & 6.71 & 6.63 & 7.30 & 7.28\\
            \rowcolor{lightgray1} \methodName{} \hspace{0.1em}\textit{Iter1} & 6.53 & 6.41 & 6.99 & 6.65 \\  
            \rowcolor{lightgray2} \methodName{} \hspace{0.1em}\textit{Iter2} & 6.66 & 6.65 & 7.34 & 7.30 \\  
            \rowcolor{lightgray3} \methodName{} \hspace{0.1em}\textit{Iter3} & \bf 6.86 & \bf 6.82 & 7.34 & \bf 7.34 \\  
            \rowcolor{lightgray4} \methodName{} \hspace{0.1em}\textit{Iter4} & 6.73 & 6.69 & \bf 7.43 & 7.04 \\  
            \bottomrule  
        \end{tabular} 
        }
        \caption{Multi-turn evaluation on MT-Bench. An asterisk (*) denotes the best score across multiple iterations. For Sampling-Ranking, Llama's best is from iteration 4 and Mistral's from iteration 3. For Self-Rewarding, both are from iteration 3.  \methodName{} progressively enhances the multi-turn instruction-following capabilities of LLMs.  
}
        \label{tab:mtbench}  
    \end{minipage}%
    \hfill  
    \begin{minipage}[b]{0.45\linewidth}  
        \centering  
        \includegraphics[width=\linewidth]{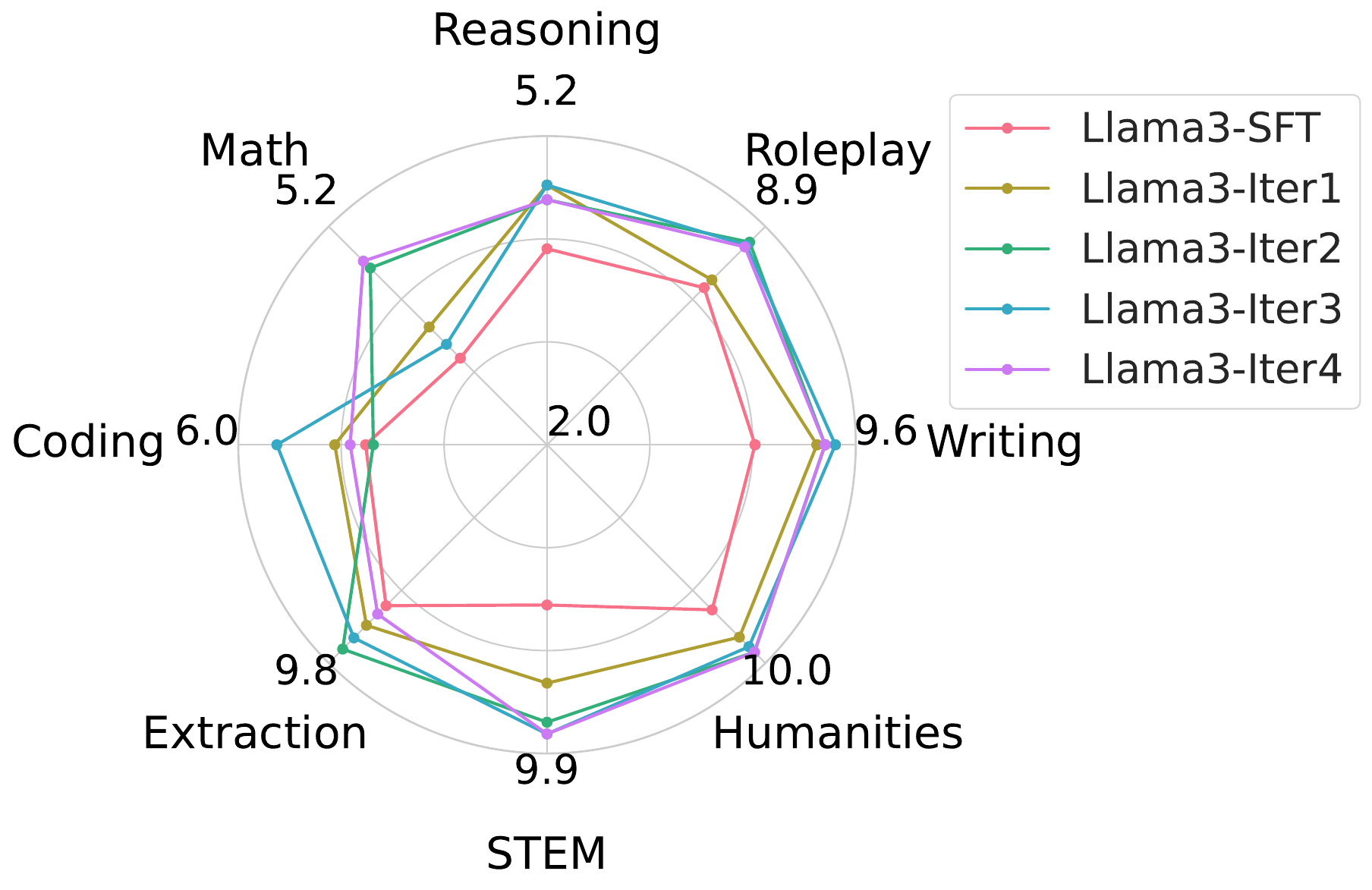}  
        \vspace{-10pt}
        \makeatletter\def\@captype{figure}\makeatother\caption{Radar chart for Llama3-8B-Base-\methodName{} on MT-Bench. \methodName{} achieves notable improvements across various prompt categories, particularly in RolePlay, STEM, Reasoning, and Coding tasks.}\label{fig:mtbench}
    \end{minipage}  
\end{table}

\subsection{Preference Alignment}
We evaluate the model alignment performance on three benchmarks: AlpacaEval 2.0~\citep{alpacaeval}, Arena-Hard~\citep{arenahard}, and MT-Bench~\citep{mtbench}.
AlpacaEval 2.0 includes 805 user prompts and utilizes pair-wise comparison with LLM-as-a-Judge. Specifically, the win rate against the baseline GPT-4 Turbo model is determined based on GPT-4 Turbo evaluation. 
Arena-Hard includes 500 more challenging user queries, employing GPT-4-Turbo to judge the model responses against GPT-4.
MT-Bench features 80 multi-turn questions spanning various domains, with GPT-4 scoring the model responses out of 10.\footnote{More details on API usage for LLM judgement are listed in Appendix~\ref{app:api} and~\ref{app:eval}.}

\paragraph{Single-Turn Dialogues}
We compare the instruction-following and human preference alignment capabilities on AlpacaEval 2.0~\citep{alpacaeval} and Arena-Hard~\citep{arenahard} in Table~\ref{tab:main_res}. Compared to the initial model post-SFT, SynPO shows sustained improvement over four iterations in win rate against GPT-4 Turbo or GPT-4. On AlpacaEval 2.0, Mistral-Base achieves a 27.4\% increase in length-controlled win rate and a 32.8\% increase in raw win rate after four iterations. Similarly, Llama3 exhibits a 26.7\% rise in length-controlled win rate and a 30.5\% improvement in raw win rate after the same number of iterations. In the more challenging Arena-Hard setting, SynPO reaches the highest win rate after the third iteration. 
Compared to the baseline methods, SynPO's iterative preference learning on synthetic data yielded more significant improvements. 

\paragraph{Multi-Turn Dialogues}
For the multi-turn benchmark MT-Bench, we report both the first-turn and second-turn scores (in Table~\ref{tab:mtbench}) as well as a radar chart depicting performance across different question types (refer to Figure~\ref{fig:mtbench}).\footnote{Similar results for Mistral-7B are shown in Figure~\ref{app:mtbenchradar} in Appendix~\ref{app:eval}.} The results indicate that SynPO enhances not only first-turn performance, with an increase of over 0.7 points, but also subsequent turns, with an increase of over 1.2 points. Compared to the initial model, SynPO shows improved performance across various question types, particularly in humanities, writing, STEM, and roleplaying.

\setlength{\tabcolsep}{3.5pt}
\begin{table}[t]
    \centering
    \resizebox{0.95\textwidth}{!}{
     \begin{tabular}{l|ccccccc}
        \toprule                
       \bf  Model    & \bf  Arc & \bf HellaSwag & \bf  TQA & \bf MMLU & \bf Winogrande & \bf GSM8k & \bf Average   \\
        \midrule
    Mistral-Base-SFT & 58.02 & 80.89 & 40.37 & 58.78 & 76.40 & 34.72 & 58.20 \\  
    \midrule
    Manual Collection & 62.71 & 83.39 & 50.69 & 58.47 & 77.35 & 32.83 & 60.91 \\
    Sampling-Ranking \hspace{0.5em}\textit{Iters*} & 60.32 & 81.80 & 44.43 & 59.09 & 76.95 & 36.85 & 59.91\\
    Self-Rewarding \hspace{0.5em}\textit{Iters*} & 60.15 & 81.84 & 43.25 & 58.98 & 76.48 & 34.72 & 59.24\\
    \rowcolor{lightgray1} Mistral-Base-\methodName{} \hspace{0.5em}\textit{Iter1}  & 60.49 & 82.25 & 50.36 & 59.00 & 76.48 & 36.39 & 60.83 \\  
    \rowcolor{lightgray2} Mistral-Base-\methodName{} \hspace{0.5em}\textit{Iter2}  & 63.65 & 83.24 & 58.04 & 58.74 & 76.48 & 27.35 & 61.25 \\  
    \rowcolor{lightgray3} Mistral-Base-\methodName{} \hspace{0.5em}\textit{Iter3}  & 63.54 & 83.14 & 58.11 & 58.37 & 75.77 & 25.26 & 60.70 \\  
    \rowcolor{lightgray4} Mistral-Base-\methodName{} \hspace{0.5em}\textit{Iter4}  & 63.57 & 83.04 & 56.12 & 58.75 & 75.77 & 31.08 & \bf 61.39 \\ 
    \midrule
    LLama3-Base-SFT  & 60.92 & 81.28 & 45.37 & 63.80 & 76.72 & 51.93 & 63.34 \\ 
    \midrule
    Manual Collection  & 66.72 & 82.89 & 59.47 & 63.10 & 77.82 & 45.72 & 65.95  \\
    Sampling-Ranking \hspace{0.5em}\textit{Iters*} & 66.38 & 82.71 & 59.84 & 63.37 & 77.27 & 54.40 & 67.33\\
    Self-Rewarding \hspace{0.5em}\textit{Iters*} & 64.76 & 82.48 & 55.54 & 63.42 & 77.03 & 54.59 & 66.30 \\
    \rowcolor{lightgray1} LLama3-Base-\methodName{} \hspace{0.5em}\textit{Iter1}  & 63.99 & 82.66 & 54.20 & 64.02 & 77.51 & 56.10 & 66.41 \\  
    \rowcolor{lightgray2} LLama3-Base-\methodName{} \hspace{0.5em}\textit{Iter2}  & 65.70 & 83.22 & 61.73 & 64.03 & 76.56 & 56.25 & 67.92 \\  
    \rowcolor{lightgray3} LLama3-Base-\methodName{} \hspace{0.5em}\textit{Iter3}  & 66.55 & 83.57 & 63.53 & 63.91 & 76.80 & 55.27 & 68.27 \\  
    \rowcolor{lightgray4} LLama3-Base-\methodName{} \hspace{0.5em}\textit{Iter4}  & 66.47 & 83.44 & 63.69 & 63.79 & 76.90 & 55.72 & \bf 68.34 \\ 
        \bottomrule
        \end{tabular} 
        }
    \caption{Open LLM Leaderboard results. TQA stands for TruthfulQA. The asterisk (*) represents the best performance across multiple iterations. Compared to the SFT versions, \methodName{} achieves an overall improvement of 3.19\% for Mistral and 5.00\% for Llama3 on the average score.}
    \label{tab:openllm}
\end{table}

\subsection{Downstream Task Performance}

In terms of the general model performance on various tasks, we report the average scores on the well-recognized Open LLM Leaderboard~\citep{open-llm-leaderboard} and 6 additional benchmarks from Language Model Evaluation Harness library (LLM Harness)~\citep{openllm}.
Open LLM Leaderboard~\citep{open-llm-leaderboard} is recognized as a standard assessment for the general performance of LLMs. It includes six different datasets, evaluating LLMs on commonsense reasoning (Arc~\citep{arc}, HellaSwag~\citep{zellers2019hellaswag}, Winogrande~\citep{sakaguchi2021winogrande}), wide knowledge (MMLU~\citep{mmlu}, TruthfulQA~\citep{lin2021truthfulqa}), and math (GSM8k~\citep{gsm8k}).
The six addtional LLM Harness tasks include Openbook Question Answering (OBQA)~\citep{obqa} and Haerae~\citep{haerae} for model knowledge, MathQA~\citep{mathqa}, XNLI~\citep{xnli}, and PROST~\citep{prost} for reasoning, as well as Toxigen~\citep{toxigen} for toxicity evaluation.\footnote{We follow the standard few-shot setting on Open LLM Leaderboard, as elaborated in Appendix~\ref{app:eval}. For the six additional tasks, we employ a fixed 5-shot setting for evaluation.}

\paragraph{Open LLM Leaderboard}
On the Open LLM leaderboard, we observe an overall improvement of 3.19\% of Mistral-Base-SFT and 5.00\% of Llama3-Base-SFT on the average score.
Specifically, SynPO achieves over 6\% improvement on the ARC challenge and over 16\% on TruthfulQA compared to the SFT model, after four rounds of self-boosting.
Notably, the performance of Mistral-Base on GSM8K has experienced a decline, whereas Llama3-Base has demonstrated an improvement of nearly 4 points on the same benchmark. 
This disparity likely stems from the superior data filtering capability of ArmoRM-Llama3-8B-v0.1 compared to the 0.4B PairRM. ArmoRM-Llama3-8B-v0.1 effectively mitigates erroneous responses and enhances mathematical problem-solving performance.

\paragraph{LLM Harness Tasks}

\setlength{\tabcolsep}{5.5pt}  
\begin{table}[t]  
    \centering  
    \resizebox{0.95\textwidth}{!}{  
    \begin{tabular}{l|ccccccc}  
        \toprule  
        \bf Model & \bf OBQA & \bf Haerae & \bf MathQA & \bf XNLI & \bf Toxigen & \bf PROST & \bf Average \\  
        \midrule  
        Mistral-Base-SFT & 46.40 & 39.96 & 36.25 & 43.76 & 60.11 & 52.04 & 46.42 \\  
        \midrule
        Manual Collection & 50.20 & 40.35 & 36.72 & 44.65 & 62.83 & 54.66 &48.24  \\
        Sampling-Ranking \hspace{0.5em}\textit{Iters*}  & 
47.60 & 40.15 & 36.21 & 44.71 & 62.45 & 53.74 & 47.48\\
        Self-Rewarding \hspace{0.5em}\textit{Iters*} &48.60 & 40.15 & 36.25 & 44.17 & 61.28 & 53.29 & 47.29

 \\
        \rowcolor{lightgray1} Mistral-Base-SynPO \hspace{0.5em}\textit{Iter1} & 48.00 & 39.78 & 36.05 & 44.23 & 60.74 & 52.39 & 46.87 \\  
        \rowcolor{lightgray2} Mistral-Base-SynPO \hspace{0.5em}\textit{Iter2} & 50.40 & 40.05 & 36.85 & 43.97 & 60.96 & 53.04 & 47.55 \\  
        \rowcolor{lightgray3} Mistral-Base-SynPO \hspace{0.5em}\textit{Iter3} & 51.20 & 40.51 & 35.88 & 44.37 & 60.74 & 53.45 & 47.69 \\  
        \rowcolor{lightgray4} Mistral-Base-SynPO \hspace{0.5em}\textit{Iter4} & 51.40 & 40.88 & 36.48 &  44.47 &  63.40 & 55.05 & \bf 48.61 \\ 
        \midrule  
        Llama3-Base-SFT & 46.20 & 61.78 & 42.04 & 45.47 & 68.83 & 52.40 & 52.79 \\  
        \midrule
    Manual Collection & 51.60& 61.59 & 42.51 & 44.77 & 74.38 & 55.96 &55.14 \\
    Sampling-Ranking \hspace{0.5em}\textit{Iters*} & 50.80 & 62.05 & 42.81 & 46.67 & 74.68 & 55.41 & 55.40\\
    Self-Rewarding \hspace{0.5em}\textit{Iters*} & 48.60 & 62.42 & 42.78 & 46.31 & 74.15 & 54.23 & 54.75\\
        \rowcolor{lightgray1} Llama3-Base-SynPO \hspace{0.5em}\textit{Iter1} & 48.20 & 61.96 & 42.71 & 46.15 & 71.17 & 53.77 & 53.99 \\  
        \rowcolor{lightgray2} Llama3-Base-SynPO \hspace{0.5em}\textit{Iter2} & 50.80 & 62.60 & 42.75 & 46.34 & 74.04 & 54.85 & 55.23 \\  
        \rowcolor{lightgray3} Llama3-Base-SynPO \hspace{0.5em}\textit{Iter3} & 51.00 & 62.51 &  42.78 &  46.37 & 75.11 & 55.29 & 55.51 \\  
        \rowcolor{lightgray4} Llama3-Base-SynPO \hspace{0.5em}\textit{Iter4} &  52.00 &  62.51 & 42.65 & 46.27 &  75.18 &  56.14 & \bf 55.79 \\  
        \bottomrule  
    \end{tabular}  
    }  
    \caption{Downstream performance in each SynPO iteration on six tasks in LM Evaluation Harness. An asterisk (*)  represents the best performance across multiple iterations.
    }  
    \label{tab:downstream2}  
\end{table}

The advantages above are also reflected in the results for more diverse tasks in LLM Harness, as evidenced by Table~\ref{tab:downstream2}.  
Previous works~\citep{sppo,meng2024simpo} demonstrate that preference optimization can induce the ``alignment tax'' - aligning models with human preferences can improve performance for only 1$\sim$2 iterations or even degrade overall performance on downstream tasks~\citep{Askell2021AGL}. Our method exhibits similar behavior on MathQA; however, overall, SynPO shows improvements across more iterations on other tasks. This is because synthesizing better chosen candidates introduces additional supervision, partially mitigating the alignment tax issue and enabling LLMs to continuously enhance their capabilities on downstream tasks at the same time of alignment.

\section{Ablation Studies} 

\subsection{Synthetic Prompts and Responses} 
\label{sec:prompt-ablt}

We have demonstrated the diversity of prompts generated by SynPO in Section~\ref{method:prompt-gen}. 
To further validate the self-prompt generator, we compare the generated prompts with manual collected prompts~\citep{cui2023ultrafeedback} and Self-Instruct prompts~\citep{Wang2022SelfInstructAL}. 
For each prompt construction approach, we randomly sample 20k prompts and construct response pairs through both SynPO and Sampling-Ranking.
We compare the single iteration results of Llama3-8B. As shown in Table~\ref{tab:prompt-ablt}, 
whether through self-refinement or Sampling-Ranking, synthetic prompts generated by SynPO lead to better-aligned models, validating the quality of these prompts. 
It is worth mentioning that SynPO prompts are even more effective than the superset of its seed training data, UltraFeedback prompts.
This increased effectiveness may be attributed to the greater diversity of SynPO prompts, achieved through the keyword sampling process in prompt synthesis. 
Results of mixing SynPO and manual collected prompts further indicate the potential of SynPO in augmenting existing prompts.

\subsection{Impact of Seed Data}
SynPO involves training LLMs solely on synthetic preference data while using seed SFT data for validation.
To investigate the maximum impact of the seed SFT data, we compare SynPO with the following settings: 1) \textit{Seed SFT:} Directly fine-tuning the LLM using seed data. 2) \textit{Seed PO:} For each prompt in the seed SFT data, using the gold standard response in the seed data as the chosen response and the initial policy model response as the rejected response for preference optimization. 3) \textit{Seed SFT + PO:} To avoid distribution shifts in directly using seed SFT data, we first obtain a model fine-tuned on seed data as in 1), then construct preference data using the model output and gold standard responses. 4) \textit{Seed SFT + PO$^{\text{me}}$:} Training on data from 3) for multiple epochs.\footnote{We compare the results of 2 epochs and 3 epochs and select the better-performing 2 epochs. More experimental details in Appendix~\ref{app:seed}} 
The results on AlpacaEval 2.0 are presented in Table~\ref{tab:seeddata}.
Among the evaluated methods except for SynPO, setting 3) is most analogous to SynPO, and proves to be the most effective. 
However, due to the limited quantity of seed data, the improvement is less than that achieved by iterative SynPO on synthetic data. Training under such conditions for multiple epochs does not yield further improvements and even degrades performance. These findings validate that SynPO is a promising approach to construct preference data and maximize the utilization of minimal high-quality data.

\begin{table}[t]  
    \centering  
    \begin{minipage}[b]{0.55\textwidth}  
        \setlength{\tabcolsep}{2pt}  
        \centering  
        \resizebox{0.99\textwidth}{!}{
        \begin{tabular}{l|cccc}  
            \toprule  
            \multirow{2}{*}{\textbf{Prompts}} & \multicolumn{2}{c}{\textbf{SynPO}} & \multicolumn{2}{c}{\textbf{Sampling-Ranking}} \\  
            \cmidrule(lr){2-3} \cmidrule(lr){4-5}  
            & \bf LC (\%) & \bf WR (\%) & \bf LC (\%) & \bf WR (\%) \\  
            \midrule  
            Manual Collection & 23.8 & 25.6 & 7.8 & 6.2 \\  
            Self-Instruct & 21.7 & 21.4 & 9.1 & 4.6 \\  
            SynPO & \bf 24.3 & 24.5 & 9.2 & 5.3 \\  
            SynPO Mix. & 23.4 & \bf 29.4 & \bf 14.2 & \bf 6.9 \\  
            \bottomrule  
        \end{tabular} 
        }
        \caption{Comparison of various prompt generation methods on AlpacaEval 2.0. SynPO Mix. combines SynPO prompts with manually collected prompts.}  
        \label{tab:prompt-ablt}  
  
    \end{minipage}  
    \hfill
    \begin{minipage}[b]{0.42\textwidth}  
        \centering  
        \setlength{\tabcolsep}{6pt}
        \resizebox{0.99\textwidth}{!}{
        \begin{tabular}{lcc}  
            \toprule  
            \bf Method & \bf LC (\%) & \bf WR (\%) \\  
            \midrule  
            Seed SFT & 20.1 & 19.7 \\  
            Seed PO & 11.6 & 10.7 \\  
            Seed SFT + PO & 24.6 & 20.9 \\  
            Seed SFT + PO$^{\text{me}}$ & 22.4 & 15.0 \\  
            SynPO \hspace{0.5em}\textit{Iter4} & \bf 32.1 & \bf 33.6 \\  
            \bottomrule  
        \end{tabular} } 
        \caption{Impact analysis of seed data on AlpacaEval 2.0. PO$^{\text{me}}$ refers to preference optimization over multiple epochs.}  
        \label{tab:seeddata}  
    \end{minipage}  
\end{table}

\section{Related Work}  

\paragraph{Preference Data Construction}
Preference data are triplets consisting of user prompts, user-preferred responses, and non-preferred responses. 
Acquiring preference data from humans can be resource intensive, often constrained by the data collection platform~\citep{ouyang2022training} or the cost of human annotation~\citep{bai2022training, pmlr-v162-ethayarajh22a, nakano2021webgpt}. 
To alleviate this problem, researchers have started using teacher LLMs, such as GPT-4~\citep{achiam2023gpt}, to simulate human preferences~\citep{cui2023ultrafeedback, ding2023enhancing, open_hermes_preferences}. Given user prompts and candidate responses, this line of work employs a stronger model to annotate preferences, thereby overcoming the scarcity and constraints of existing preference data~\citep{cui2023ultrafeedback}. 
However, the prompts still need to be collected~\citep{cui2023ultrafeedback}, which limits the domain, diversity, and quantity of the data. Both human annotation and the utilization of large teacher model APIs incur substantial costs~\citep{Shi2023SaferInstructAL}. 
Moreover, relying solely on reward scores or win-lose annotation fails to fully capture the subtleties and complexities of human preferences.

\paragraph{LLM Self-Boosting}
Previous work has advanced the self-boosting of LLMs by searching for high-reward behaviors~\citep{Tian2024TowardSO,Zhang2024ReSTMCTSLS}, using LLMs as judges to select responses~\citep{Yuan2024SelfRewardingLM,Wang2024SelfTaughtE,Wu2024MetaRewardingLM,Kim2024AligningLL}, and leveraging self-play strategies~\citep{spin,Cheng2024SelfplayingAL,Luo2024ArenaLB,sppo}. 
These works typically use a fixed set of existing prompts, limiting the LLM ability to learn across wide scenarios.
Furthermore, the deterministic reward signals in these methods do not help the model recognize subtle discrepancies between its responses and ideal responses.
Prior to our work, Constitutional AI~\citep{Bai2022ConstitutionalAH} and SELF~\citep{jqself} used AI to generate non-deterministic feedback for training. Constitutional AI used refinement data for reward models, while SELF employed GPT-4 for data generation and taught models self-refinement. However, these methods did not utilize comparative information between pre- and post-revision texts for training.

\paragraph{Synthetic Data for LLMs}
Acquiring human-generated data is costly and time-consuming, leading to the use of synthetic data for LLM training~\citep{Wang2022SelfInstructAL,xu2023wizardlm,Li2024SyntheticD}. Unnatural Instructions~\citep{Honovich2022UnnaturalIT} and Self-Instruct~\citep{Wang2022SelfInstructAL} use seed instructions to generate new prompts, while WizardLM~\citep{xu2023wizardlm} and WizardMath~\citep{luo2023wizardmath} rewrite these instructions into more complex forms using ChatGPT. Seed topics also produce textbook-like data~\citep{phi15,Li2024SyntheticD} or self-chat dialogues~\citep{xu2023baize,ding2023enhancing} for instruction tuning. These methods often require strong LLMs or examples, benefiting from model distillation~\citep{xu2023wizardlm,Li2024SyntheticD,Li2024HRLAIFII}.
Our approach uses the model itself to generate prompts and responses without needing carefully designed topics, extending beyond the model's inherent sampling space. This brings generative rewards to the LLM self-boosting process, particularly benefiting initially weaker models. It resembles direct preference knowledge distillation~\citep{dpkd} but does not rely on large-scale teacher model responses for supervision.

\section{Conclusion}
We introduce self-boosting LLM with synthetic preference data, SynPO, a method for LLM alignment through iterative training on synthetic data. 
In SynPO, we innovatively base the entire training process on synthetic data and only employ limited SFT data for validation. 
SynPO diversifies the prompts and dynamically guides LLMs to improve their own output, using pre- and post-refinement generations as synthetic preference pairs for training in the next iteration.
Experimental results show that SynPO leads to significant improvements on both instruction-following capabilities and task performance. This strategy sheds light on high-quality synthetic data generation and self-alignment with minimal supervision, both of which are critical for the continuous development of LLMs.

\bibliography{iclr2025_conference}

\begin{thebibliography}{68}
\providecommand{\natexlab}[1]{#1}
\providecommand{\url}[1]{\texttt{#1}}
\expandafter\ifx\csname urlstyle\endcsname\relax
  \providecommand{\doi}[1]{doi: #1}\else
  \providecommand{\doi}{doi: \begingroup \urlstyle{rm}\Url}\fi

\bibitem[Achiam et~al.(2023)Achiam, Adler, Agarwal, Ahmad, Akkaya, Aleman, Almeida, Altenschmidt, Altman, Anadkat, et~al.]{achiam2023gpt}
Josh Achiam, Steven Adler, Sandhini Agarwal, Lama Ahmad, Ilge Akkaya, Florencia~Leoni Aleman, Diogo Almeida, Janko Altenschmidt, Sam Altman, Shyamal Anadkat, et~al.
\newblock Gpt-4 technical report.
\newblock \emph{arXiv preprint arXiv:2303.08774}, 2023.

\bibitem[Amini et~al.(2019)Amini, Gabriel, Lin, Koncel-Kedziorski, Choi, and Hajishirzi]{mathqa}
Aida Amini, Saadia Gabriel, Peter Lin, Rik Koncel-Kedziorski, Yejin Choi, and Hannaneh Hajishirzi.
\newblock Mathqa: Towards interpretable math word problem solving with operation-based formalisms, 2019.

\bibitem[Aroca-Ouellette et~al.(2021)Aroca-Ouellette, Paik, Roncone, and Kann]{prost}
St{\'e}phane Aroca-Ouellette, Cory Paik, Alessandro Roncone, and Katharina Kann.
\newblock {PROST}: {P}hysical reasoning about objects through space and time.
\newblock In \emph{Findings of the Association for Computational Linguistics: ACL-IJCNLP 2021}, pp.\  4597--4608, Online, August 2021. Association for Computational Linguistics.
\newblock URL \url{https://aclanthology.org/2021.findings-acl.404}.

\bibitem[Askell et~al.(2021)Askell, Bai, Chen, Drain, Ganguli, Henighan, Jones, Joseph, Mann, Dassarma, Elhage, Hatfield-Dodds, Hernandez, Kernion, Ndousse, Olsson, Amodei, Brown, Clark, McCandlish, Olah, and Kaplan]{Askell2021AGL}
Amanda Askell, Yuntao Bai, Anna Chen, Dawn Drain, Deep Ganguli, Tom Henighan, Andy Jones, Nicholas Joseph, Benjamin Mann, Nova Dassarma, Nelson Elhage, Zac Hatfield-Dodds, Danny Hernandez, John Kernion, Kamal Ndousse, Catherine Olsson, Dario Amodei, Tom~B. Brown, Jack Clark, Sam McCandlish, Christopher Olah, and Jared Kaplan.
\newblock A general language assistant as a laboratory for alignment.
\newblock \emph{ArXiv}, abs/2112.00861, 2021.
\newblock URL \url{https://api.semanticscholar.org/CorpusID:244799619}.

\bibitem[Bai et~al.(2022{\natexlab{a}})Bai, Jones, Ndousse, Askell, Chen, DasSarma, Drain, Fort, Ganguli, Henighan, Joseph, Kadavath, Kernion, Conerly, El-Showk, Elhage, Hatfield-Dodds, Hernandez, Hume, Johnston, Kravec, Lovitt, Nanda, Olsson, Amodei, Brown, Clark, McCandlish, Olah, Mann, and Kaplan]{bai2022training}
Yuntao Bai, Andy Jones, Kamal Ndousse, Amanda Askell, Anna Chen, Nova DasSarma, Dawn Drain, Stanislav Fort, Deep Ganguli, Tom Henighan, Nicholas Joseph, Saurav Kadavath, Jackson Kernion, Tom Conerly, Sheer El-Showk, Nelson Elhage, Zac Hatfield-Dodds, Danny Hernandez, Tristan Hume, Scott Johnston, Shauna Kravec, Liane Lovitt, Neel Nanda, Catherine Olsson, Dario Amodei, Tom Brown, Jack Clark, Sam McCandlish, Chris Olah, Ben Mann, and Jared Kaplan.
\newblock Training a helpful and harmless assistant with reinforcement learning from human feedback, 2022{\natexlab{a}}.

\bibitem[Bai et~al.(2022{\natexlab{b}})Bai, Kadavath, Kundu, Askell, Kernion, Jones, Chen, Goldie, Mirhoseini, McKinnon, Chen, Olsson, Olah, Hernandez, Drain, Ganguli, Li, Tran-Johnson, Perez, Kerr, Mueller, Ladish, Landau, Ndousse, Lukoiūtė, Lovitt, Sellitto, Elhage, Schiefer, Mercado, Dassarma, Lasenby, Larson, Ringer, Johnston, Kravec, Showk, Fort, Lanham, Telleen-Lawton, Conerly, Henighan, Hume, Bowman, Hatfield-Dodds, Mann, Amodei, Joseph, McCandlish, Brown, and Kaplan]{Bai2022ConstitutionalAH}
Yuntao Bai, Saurav Kadavath, Sandipan Kundu, Amanda Askell, John Kernion, Andy Jones, Anna Chen, Anna Goldie, Azalia Mirhoseini, Cameron McKinnon, Carol Chen, Catherine Olsson, Christopher Olah, Danny Hernandez, Dawn Drain, Deep Ganguli, Dustin Li, Eli Tran-Johnson, E~Perez, Jamie Kerr, Jared Mueller, Jeff Ladish, J~Landau, Kamal Ndousse, Kamilė Lukoiūtė, Liane Lovitt, Michael Sellitto, Nelson Elhage, Nicholas Schiefer, Noem'i Mercado, Nova Dassarma, Robert Lasenby, Robin Larson, Sam Ringer, Scott Johnston, Shauna Kravec, Sheer~El Showk, Stanislav Fort, Tamera Lanham, Timothy Telleen-Lawton, Tom Conerly, Tom Henighan, Tristan Hume, Sam Bowman, Zac Hatfield-Dodds, Benjamin Mann, Dario Amodei, Nicholas Joseph, Sam McCandlish, Tom~B. Brown, and Jared Kaplan.
\newblock Constitutional ai: Harmlessness from ai feedback.
\newblock \emph{ArXiv}, abs/2212.08073, 2022{\natexlab{b}}.
\newblock URL \url{https://api.semanticscholar.org/CorpusID:254823489}.

\bibitem[Beeching et~al.(2023)Beeching, Fourrier, Habib, Han, Lambert, Rajani, Sanseviero, Tunstall, and Wolf]{open-llm-leaderboard}
Edward Beeching, Clémentine Fourrier, Nathan Habib, Sheon Han, Nathan Lambert, Nazneen Rajani, Omar Sanseviero, Lewis Tunstall, and Thomas Wolf.
\newblock Open llm leaderboard.
\newblock \url{https://huggingface.co/spaces/open-llm-leaderboard-old/open_llm_leaderboard}, 2023.

\bibitem[Chen et~al.(2024)Chen, Deng, Yuan, Ji, and Gu]{spin}
Zixiang Chen, Yihe Deng, Huizhuo Yuan, Kaixuan Ji, and Quanquan Gu.
\newblock Self-play fine-tuning converts weak language models to strong language models.
\newblock \emph{arXiv preprint arXiv:2401.01335}, 2024.

\bibitem[Cheng et~al.(2024)Cheng, Hu, Xu, Zhang, Dai, Han, and Du]{Cheng2024SelfplayingAL}
Pengyu Cheng, Tianhao Hu, Han Xu, Zhisong Zhang, Yong Dai, Lei Han, and Nan Du.
\newblock Self-playing adversarial language game enhances llm reasoning.
\newblock \emph{ArXiv}, abs/2404.10642, 2024.
\newblock URL \url{https://api.semanticscholar.org/CorpusID:269157364}.

\bibitem[Chiang et~al.(2023)Chiang, Li, Lin, Sheng, Wu, Zhang, Zheng, Zhuang, Zhuang, Gonzalez, Stoica, and Xing]{vicuna2023}
Wei-Lin Chiang, Zhuohan Li, Zi~Lin, Ying Sheng, Zhanghao Wu, Hao Zhang, Lianmin Zheng, Siyuan Zhuang, Yonghao Zhuang, Joseph~E. Gonzalez, Ion Stoica, and Eric~P. Xing.
\newblock Vicuna: An open-source chatbot impressing gpt-4 with 90\%* chatgpt quality, March 2023.
\newblock URL \url{https://lmsys.org/blog/2023-03-30-vicuna/}.

\bibitem[Clark et~al.(2018)Clark, Cowhey, Etzioni, Khot, Sabharwal, Schoenick, and Tafjord]{arc}
Peter Clark, Isaac Cowhey, Oren Etzioni, Tushar Khot, Ashish Sabharwal, Carissa Schoenick, and Oyvind Tafjord.
\newblock Think you have solved question answering? try arc, the ai2 reasoning challenge.
\newblock \emph{arXiv preprint arXiv:1803.05457}, 2018.

\bibitem[Cobbe et~al.(2021)Cobbe, Kosaraju, Bavarian, Chen, Jun, Kaiser, Plappert, Tworek, Hilton, Nakano, et~al.]{gsm8k}
Karl Cobbe, Vineet Kosaraju, Mohammad Bavarian, Mark Chen, Heewoo Jun, Lukasz Kaiser, Matthias Plappert, Jerry Tworek, Jacob Hilton, Reiichiro Nakano, et~al.
\newblock Training verifiers to solve math word problems.
\newblock \emph{arXiv preprint arXiv:2110.14168}, 2021.

\bibitem[Conneau et~al.(2018)Conneau, Rinott, Lample, Williams, Bowman, Schwenk, and Stoyanov]{xnli}
Alexis Conneau, Ruty Rinott, Guillaume Lample, Adina Williams, Samuel~R. Bowman, Holger Schwenk, and Veselin Stoyanov.
\newblock Xnli: Evaluating cross-lingual sentence representations.
\newblock In \emph{Proceedings of the 2018 Conference on Empirical Methods in Natural Language Processing}. Association for Computational Linguistics, 2018.

\bibitem[Cui et~al.(2023)Cui, Yuan, Ding, Yao, Zhu, Ni, Xie, Liu, and Sun]{cui2023ultrafeedback}
Ganqu Cui, Lifan Yuan, Ning Ding, Guanming Yao, Wei Zhu, Yuan Ni, Guotong Xie, Zhiyuan Liu, and Maosong Sun.
\newblock Ultrafeedback: Boosting language models with high-quality feedback, 2023.

\bibitem[Ding et~al.(2023)Ding, Chen, Xu, Qin, Zheng, Hu, Liu, Sun, and Zhou]{ding2023enhancing}
Ning Ding, Yulin Chen, Bokai Xu, Yujia Qin, Zhi Zheng, Shengding Hu, Zhiyuan Liu, Maosong Sun, and Bowen Zhou.
\newblock Enhancing chat language models by scaling high-quality instructional conversations, 2023.

\bibitem[Dubey et~al.(2024)Dubey, Jauhri, Pandey, Kadian, Al-Dahle, Letman, Mathur, Schelten, Yang, Fan, et~al.]{Llama31}
Abhimanyu Dubey, Abhinav Jauhri, Abhinav Pandey, Abhishek Kadian, Ahmad Al-Dahle, Aiesha Letman, Akhil Mathur, Alan Schelten, Amy Yang, Angela Fan, et~al.
\newblock The llama 3 herd of models.
\newblock \emph{arXiv preprint arXiv:2407.21783}, 2024.

\bibitem[Dubois et~al.(2024)Dubois, Galambosi, Liang, and Hashimoto]{alpacaeval}
Yann Dubois, Bal{\'a}zs Galambosi, Percy Liang, and Tatsunori~B Hashimoto.
\newblock Length-controlled alpacaeval: A simple way to debias automatic evaluators.
\newblock \emph{arXiv preprint arXiv:2404.04475}, 2024.

\bibitem[Ethayarajh et~al.(2022)Ethayarajh, Choi, and Swayamdipta]{pmlr-v162-ethayarajh22a}
Kawin Ethayarajh, Yejin Choi, and Swabha Swayamdipta.
\newblock Understanding dataset difficulty with $\mathcal{V}$-usable information.
\newblock In Kamalika Chaudhuri, Stefanie Jegelka, Le~Song, Csaba Szepesvari, Gang Niu, and Sivan Sabato (eds.), \emph{Proceedings of the 39th International Conference on Machine Learning}, volume 162 of \emph{Proceedings of Machine Learning Research}, pp.\  5988--6008. PMLR, 2022.

\bibitem[Ethayarajh et~al.(2024)Ethayarajh, Xu, Muennighoff, Jurafsky, and Kiela]{kto}
Kawin Ethayarajh, Winnie Xu, Niklas Muennighoff, Dan Jurafsky, and Douwe Kiela.
\newblock Kto: Model alignment as prospect theoretic optimization.
\newblock \emph{ArXiv}, abs/2402.01306, 2024.
\newblock URL \url{https://api.semanticscholar.org/CorpusID:267406810}.

\bibitem[Gao et~al.(2024)Gao, Tow, Abbasi, Biderman, Black, DiPofi, Foster, Golding, Hsu, Le~Noac'h, Li, McDonell, Muennighoff, Ociepa, Phang, Reynolds, Schoelkopf, Skowron, Sutawika, Tang, Thite, Wang, Wang, and Zou]{openllm}
Leo Gao, Jonathan Tow, Baber Abbasi, Stella Biderman, Sid Black, Anthony DiPofi, Charles Foster, Laurence Golding, Jeffrey Hsu, Alain Le~Noac'h, Haonan Li, Kyle McDonell, Niklas Muennighoff, Chris Ociepa, Jason Phang, Laria Reynolds, Hailey Schoelkopf, Aviya Skowron, Lintang Sutawika, Eric Tang, Anish Thite, Ben Wang, Kevin Wang, and Andy Zou.
\newblock A framework for few-shot language model evaluation, 07 2024.
\newblock URL \url{https://zenodo.org/records/12608602}.

\bibitem[Gulcehre et~al.(2023)Gulcehre, Paine, Srinivasan, Konyushkova, Weerts, Sharma, Siddhant, Ahern, Wang, Gu, Macherey, Doucet, Firat, and de~Freitas]{rest}
Caglar Gulcehre, Tom~Le Paine, Srivatsan Srinivasan, Ksenia Konyushkova, Lotte Weerts, Abhishek Sharma, Aditya Siddhant, Alexa Ahern, Miaosen Wang, Chenjie Gu, Wolfgang Macherey, A.~Doucet, Orhan Firat, and Nando de~Freitas.
\newblock Reinforced self-training (rest) for language modeling.
\newblock \emph{ArXiv}, abs/2308.08998, 2023.
\newblock URL \url{https://api.semanticscholar.org/CorpusID:261031028}.

\bibitem[Hartvigsen et~al.(2022)Hartvigsen, Gabriel, Palangi, Sap, Ray, and Kamar]{toxigen}
Thomas Hartvigsen, Saadia Gabriel, Hamid Palangi, Maarten Sap, Dipankar Ray, and Ece Kamar.
\newblock Toxigen: A large-scale machine-generated dataset for implicit and adversarial hate speech detection.
\newblock In \emph{Proceedings of the 60th Annual Meeting of the Association for Computational Linguistics}, 2022.

\bibitem[Hendrycks et~al.(2020)Hendrycks, Burns, Basart, Zou, Mazeika, Song, and Steinhardt]{mmlu}
Dan Hendrycks, Collin Burns, Steven Basart, Andy Zou, Mantas Mazeika, Dawn Song, and Jacob Steinhardt.
\newblock Measuring massive multitask language understanding.
\newblock \emph{arXiv preprint arXiv:2009.03300}, 2020.

\bibitem[Honovich et~al.(2022)Honovich, Scialom, Levy, and Schick]{Honovich2022UnnaturalIT}
Or~Honovich, Thomas Scialom, Omer Levy, and Timo Schick.
\newblock Unnatural instructions: Tuning language models with (almost) no human labor.
\newblock \emph{ArXiv}, abs/2212.09689, 2022.
\newblock URL \url{https://api.semanticscholar.org/CorpusID:254853659}.

\bibitem[Hu et~al.(2023)Hu, Luo, Wang, Cheng, Liu, and Sun]{Hu2023WontGF}
Shengding Hu, Yi-Xiao Luo, Huadong Wang, Xingyi Cheng, Zhiyuan Liu, and Maosong Sun.
\newblock Won’t get fooled again: Answering questions with false premises.
\newblock In \emph{Annual Meeting of the Association for Computational Linguistics}, 2023.
\newblock URL \url{https://api.semanticscholar.org/CorpusID:259341789}.

\bibitem[Hu et~al.(2024)Hu, Li, Sheng, Chen, Chen, Mei, Ye, Zhang, and Liu]{Hu2024TowardsCP}
Yulan Hu, Qingyang Li, Ouyang Sheng, Ge~Chen, Kaihui Chen, Lijun Mei, Xucheng Ye, Fuzheng Zhang, and Yong Liu.
\newblock Towards comprehensive preference data collection for reward modeling.
\newblock \emph{ArXiv}, abs/2406.16486, 2024.
\newblock URL \url{https://api.semanticscholar.org/CorpusID:270703662}.

\bibitem[Huang et~al.(2024)Huang, Piqueres, Rasul, Schmid, Vila, and Tunstall]{open_hermes_preferences}
Shengyi~Costa Huang, Agustín Piqueres, Kashif Rasul, Philipp Schmid, Daniel Vila, and Lewis Tunstall.
\newblock Open hermes preferences.
\newblock \url{https://huggingface.co/datasets/argilla/OpenHermesPreferences}, 2024.

\bibitem[Jiang et~al.(2023)Jiang, Ren, and Lin]{pairrm}
Dongfu Jiang, Xiang Ren, and Bill~Yuchen Lin.
\newblock Llm-blender: Ensembling large language models with pairwise ranking and generative fusion.
\newblock In \emph{Annual Meeting of the Association for Computational Linguistics}, 2023.
\newblock URL \url{https://api.semanticscholar.org/CorpusID:259075564}.

\bibitem[Kim et~al.(2024)Kim, Lee, Shin, and Kim]{Kim2024AligningLL}
Dongyoung Kim, Kimin Lee, Jinwoo Shin, and Jaehyung Kim.
\newblock Aligning large language models with self-generated preference data.
\newblock \emph{ArXiv}, abs/2406.04412, 2024.
\newblock URL \url{https://api.semanticscholar.org/CorpusID:270357971}.

\bibitem[Li et~al.(2024{\natexlab{a}})Li, Xiao, Cao, Tang, Yuan, Zhao, Chen, Zhang, Li, Yang, Guo, Gan, Yu, Wang, and Shan]{Li2024HRLAIFII}
Ang Li, Qiugen Xiao, Peng Cao, Jian Tang, Yi~Yuan, Zijie Zhao, Xiaoyuan Chen, Liang Zhang, Xiangyang Li, Kaitong Yang, Weidong Guo, Yukang Gan, Jeffrey~Xu Yu, Dan~Tong Wang, and Ying Shan.
\newblock Hrlaif: Improvements in helpfulness and harmlessness in open-domain reinforcement learning from ai feedback.
\newblock \emph{ArXiv}, abs/2403.08309, 2024{\natexlab{a}}.
\newblock URL \url{https://api.semanticscholar.org/CorpusID:268379328}.

\bibitem[Li et~al.(2024{\natexlab{b}})Li, Dong, Tang, Wang, Zhang, Huang, Huang, Huang, Huang, Zhang, Gu, Cheng, Wang, Chen, Dong, Lu, Sui, Wang, Lam, and Wei]{Li2024SyntheticD}
Haoran Li, Qingxiu Dong, Zhengyang Tang, Chaojun Wang, Xingxing Zhang, Haoyang Huang, Shaohan Huang, Xiaolong Huang, Zeqiang Huang, Dongdong Zhang, Yuxian Gu, Xin Cheng, Xun Wang, Si-Qing Chen, Li~Dong, Wei Lu, Zhifang Sui, Benyou Wang, Wai Lam, and Furu Wei.
\newblock Synthetic data (almost) from scratch: Generalized instruction tuning for language models.
\newblock \emph{ArXiv}, abs/2402.13064, 2024{\natexlab{b}}.
\newblock URL \url{https://api.semanticscholar.org/CorpusID:267759981}.

\bibitem[Li et~al.(2024{\natexlab{c}})Li, Chiang, Frick, Dunlap, Wu, Zhu, Gonzalez, and Stoica]{arenahard}
Tianle Li, Wei-Lin Chiang, Evan Frick, Lisa Dunlap, Tianhao Wu, Banghua Zhu, Joseph~E. Gonzalez, and Ion Stoica.
\newblock From crowdsourced data to high-quality benchmarks: Arena-hard and benchbuilder pipeline.
\newblock \emph{ArXiv}, abs/2406.11939, 2024{\natexlab{c}}.
\newblock URL \url{https://api.semanticscholar.org/CorpusID:270562889}.

\bibitem[Li et~al.(2024{\natexlab{d}})Li, Gu, Dong, Wang, Cheng, and Wei]{dpkd}
Yixing Li, Yuxian Gu, Li~Dong, Dequan Wang, Yu~Cheng, and Furu Wei.
\newblock Direct preference knowledge distillation for large language models.
\newblock \emph{ArXiv}, abs/2406.19774, 2024{\natexlab{d}}.
\newblock URL \url{https://api.semanticscholar.org/CorpusID:270845775}.

\bibitem[Li et~al.(2023)Li, Bubeck, Eldan, Del~Giorno, Gunasekar, and Lee]{phi15}
Yuanzhi Li, S{\'e}bastien Bubeck, Ronen Eldan, Allie Del~Giorno, Suriya Gunasekar, and Yin~Tat Lee.
\newblock Textbooks are all you need ii: phi-1.5 technical report.
\newblock \emph{arXiv preprint arXiv:2309.05463}, 2023.

\bibitem[Lin et~al.(2021)Lin, Hilton, and Evans]{lin2021truthfulqa}
Stephanie Lin, Jacob Hilton, and Owain Evans.
\newblock Truthfulqa: Measuring how models mimic human falsehoods.
\newblock \emph{arXiv preprint arXiv:2109.07958}, 2021.

\bibitem[Lu et~al.(2023)Lu, Zhong, Huang, Wang, Mi, Wang, Wang, Shang, and Liu]{jqself}
Jianqiao Lu, Wanjun Zhong, Wenyong Huang, Yufei Wang, Fei Mi, Baojun Wang, Weichao Wang, Lifeng Shang, and Qun Liu.
\newblock Self: Self-evolution with language feedback.
\newblock 2023.
\newblock URL \url{https://api.semanticscholar.org/CorpusID:263334155}.

\bibitem[Luo et~al.(2023)Luo, Sun, Xu, Zhao, Lou, Tao, Geng, Lin, Chen, and Zhang]{luo2023wizardmath}
Haipeng Luo, Qingfeng Sun, Can Xu, Pu~Zhao, Jianguang Lou, Chongyang Tao, Xiubo Geng, Qingwei Lin, Shifeng Chen, and Dongmei Zhang.
\newblock Wizardmath: Empowering mathematical reasoning for large language models via reinforced evol-instruct.
\newblock \emph{arXiv preprint arXiv:2308.09583}, 2023.

\bibitem[Luo et~al.(2024)Luo, Sun, Xu, Zhao, Lin, Lou, Chen, Tang, and Chen]{Luo2024ArenaLB}
Haipeng Luo, Qingfeng Sun, Can Xu, Pu~Zhao, Qingwei Lin, Jianguang Lou, Shifeng Chen, Yansong Tang, and Weizhu Chen.
\newblock Arena learning: Build data flywheel for llms post-training via simulated chatbot arena.
\newblock \emph{ArXiv}, abs/2407.10627, 2024.
\newblock URL \url{https://api.semanticscholar.org/CorpusID:271213086}.

\bibitem[Madaan et~al.(2023)Madaan, Tandon, Gupta, Hallinan, Gao, Wiegreffe, Alon, Dziri, Prabhumoye, Yang, Welleck, Majumder, Gupta, Yazdanbakhsh, and Clark]{Madaan2023SelfRefineIR}
Aman Madaan, Niket Tandon, Prakhar Gupta, Skyler Hallinan, Luyu Gao, Sarah Wiegreffe, Uri Alon, Nouha Dziri, Shrimai Prabhumoye, Yiming Yang, Sean Welleck, Bodhisattwa~Prasad Majumder, Shashank Gupta, Amir Yazdanbakhsh, and Peter Clark.
\newblock Self-refine: Iterative refinement with self-feedback.
\newblock \emph{ArXiv}, abs/2303.17651, 2023.
\newblock URL \url{https://api.semanticscholar.org/CorpusID:257900871}.

\bibitem[Meng et~al.(2024)Meng, Xia, and Chen]{meng2024simpo}
Yu~Meng, Mengzhou Xia, and Danqi Chen.
\newblock Simpo: Simple preference optimization with a reference-free reward.
\newblock \emph{arXiv preprint arXiv:2405.14734}, 2024.

\bibitem[Mihaylov et~al.(2018)Mihaylov, Clark, Khot, and Sabharwal]{obqa}
Todor Mihaylov, Peter Clark, Tushar Khot, and Ashish Sabharwal.
\newblock Can a suit of armor conduct electricity? a new dataset for open book question answering.
\newblock In \emph{EMNLP}, 2018.

\bibitem[Nakano et~al.(2021)Nakano, Hilton, Balaji, Wu, Ouyang, Kim, Hesse, Jain, Kosaraju, Saunders, Jiang, Cobbe, Eloundou, Krueger, Button, Knight, Chess, and Schulman]{nakano2021webgpt}
Reiichiro Nakano, Jacob Hilton, Suchir Balaji, Jeff Wu, Long Ouyang, Christina Kim, Christopher Hesse, Shantanu Jain, Vineet Kosaraju, William Saunders, Xu~Jiang, Karl Cobbe, Tyna Eloundou, Gretchen Krueger, Kevin Button, Matthew Knight, Benjamin Chess, and John Schulman.
\newblock Webgpt: Browser-assisted question-answering with human feedback.
\newblock In \emph{arXiv}, 2021.

\bibitem[Nguyen et~al.(2024)Nguyen, Li, Oh, Schmidt, Weston, Zettlemoyer, and Li]{Nguyen2024BetterAW}
Thao Nguyen, Jeffrey Li, Sewoong Oh, Ludwig Schmidt, Jason Weston, Luke Zettlemoyer, and Xian Li.
\newblock Better alignment with instruction back-and-forth translation.
\newblock 2024.
\newblock URL \url{https://api.semanticscholar.org/CorpusID:271768943}.

\bibitem[Ouyang et~al.(2022{\natexlab{a}})Ouyang, Wu, Jiang, Almeida, Wainwright, Mishkin, Zhang, Agarwal, Slama, Ray, Schulman, Hilton, Kelton, Miller, Simens, Askell, Welinder, Christiano, Leike, and Lowe]{ouyang2022training}
Long Ouyang, Jeff Wu, Xu~Jiang, Diogo Almeida, Carroll~L. Wainwright, Pamela Mishkin, Chong Zhang, Sandhini Agarwal, Katarina Slama, Alex Ray, John Schulman, Jacob Hilton, Fraser Kelton, Luke Miller, Maddie Simens, Amanda Askell, Peter Welinder, Paul Christiano, Jan Leike, and Ryan Lowe.
\newblock Training language models to follow instructions with human feedback, 2022{\natexlab{a}}.

\bibitem[Ouyang et~al.(2022{\natexlab{b}})Ouyang, Wu, Jiang, Almeida, Wainwright, Mishkin, Zhang, Agarwal, Slama, Ray, Schulman, Hilton, Kelton, Miller, Simens, Askell, Welinder, Christiano, Leike, and Lowe]{Ouyang2022TrainingLM}
Long Ouyang, Jeff Wu, Xu~Jiang, Diogo Almeida, Carroll~L. Wainwright, Pamela Mishkin, Chong Zhang, Sandhini Agarwal, Katarina Slama, Alex Ray, John Schulman, Jacob Hilton, Fraser Kelton, Luke~E. Miller, Maddie Simens, Amanda Askell, Peter Welinder, Paul~Francis Christiano, Jan Leike, and Ryan~J. Lowe.
\newblock Training language models to follow instructions with human feedback.
\newblock \emph{ArXiv}, abs/2203.02155, 2022{\natexlab{b}}.
\newblock URL \url{https://api.semanticscholar.org/CorpusID:246426909}.

\bibitem[Penedo et~al.(2023)Penedo, Malartic, Hesslow, Cojocaru, Cappelli, Alobeidli, Pannier, Almazrouei, and Launay]{refinedweb}
Guilherme Penedo, Quentin Malartic, Daniel Hesslow, Ruxandra Cojocaru, Alessandro Cappelli, Hamza Alobeidli, Baptiste Pannier, Ebtesam Almazrouei, and Julien Launay.
\newblock The {R}efined{W}eb dataset for {F}alcon {LLM}: outperforming curated corpora with web data, and web data only.
\newblock \emph{arXiv preprint arXiv:2306.01116}, 2023.

\bibitem[Rafailov et~al.(2024)Rafailov, Sharma, Mitchell, Manning, Ermon, and Finn]{dpo}
Rafael Rafailov, Archit Sharma, Eric Mitchell, Christopher~D Manning, Stefano Ermon, and Chelsea Finn.
\newblock Direct preference optimization: Your language model is secretly a reward model.
\newblock \emph{Advances in Neural Information Processing Systems}, 36, 2024.

\bibitem[Reimers \& Gurevych(2019)Reimers and Gurevych]{reimers-2019-sentence-bert}
Nils Reimers and Iryna Gurevych.
\newblock Sentence-bert: Sentence embeddings using siamese bert-networks.
\newblock In \emph{Proceedings of the 2019 Conference on Empirical Methods in Natural Language Processing}. Association for Computational Linguistics, 11 2019.
\newblock URL \url{https://arxiv.org/abs/1908.10084}.

\bibitem[Rosset et~al.(2024)Rosset, Cheng, Mitra, Santacroce, Awadallah, and Xie]{Rosset2024DirectNO}
Corby Rosset, Ching-An Cheng, Arindam Mitra, Michael Santacroce, Ahmed Awadallah, and Tengyang Xie.
\newblock Direct nash optimization: Teaching language models to self-improve with general preferences.
\newblock \emph{ArXiv}, abs/2404.03715, 2024.
\newblock URL \url{https://api.semanticscholar.org/CorpusID:268987488}.

\bibitem[Sakaguchi et~al.(2021)Sakaguchi, Bras, Bhagavatula, and Choi]{sakaguchi2021winogrande}
Keisuke Sakaguchi, Ronan~Le Bras, Chandra Bhagavatula, and Yejin Choi.
\newblock Winogrande: An adversarial winograd schema challenge at scale.
\newblock \emph{Communications of the ACM}, 64\penalty0 (9):\penalty0 99--106, 2021.

\bibitem[Shi et~al.(2023)Shi, Chen, and Zhao]{Shi2023SaferInstructAL}
Taiwei Shi, Kai Chen, and Jieyu Zhao.
\newblock Safer-instruct: Aligning language models with automated preference data.
\newblock \emph{ArXiv}, abs/2311.08685, 2023.
\newblock URL \url{https://api.semanticscholar.org/CorpusID:265212888}.

\bibitem[Singh et~al.(2022)Singh, Morris, Aneja, Rush, and Gao]{singh2022explaining}
Chandan Singh, John~Xavier Morris, Jyoti Aneja, Alexander~M Rush, and Jianfeng Gao.
\newblock Explaining patterns in data with language models via interpretable autoprompting.
\newblock 2022.

\bibitem[Son et~al.(2023)Son, Lee, Kim, Kim, Lee, Yeom, Jung, Kim, and Kim]{haerae}
Guijin Son, Hanwool Lee, Suwan Kim, Huiseo Kim, Jaecheol Lee, Je~Won Yeom, Jihyu Jung, Jung~Woo Kim, and Songseong Kim.
\newblock Hae-rae bench: Evaluation of korean knowledge in language models, 2023.

\bibitem[Song et~al.(2024)Song, Yu, Lang, Yu, Huang, Wang, and Li]{Song2024ScalingDD}
Feifan Song, Bowen Yu, Hao Lang, Haiyang Yu, Fei Huang, Houfeng Wang, and Yongbin Li.
\newblock Scaling data diversity for fine-tuning language models in human alignment.
\newblock In \emph{International Conference on Language Resources and Evaluation}, 2024.
\newblock URL \url{https://api.semanticscholar.org/CorpusID:268512826}.

\bibitem[Tian et~al.(2024)Tian, Peng, Song, Jin, Yu, Mi, and Yu]{Tian2024TowardSO}
Ye~Tian, Baolin Peng, Linfeng Song, Lifeng Jin, Dian Yu, Haitao Mi, and Dong Yu.
\newblock Toward self-improvement of llms via imagination, searching, and criticizing.
\newblock \emph{ArXiv}, abs/2404.12253, 2024.
\newblock URL \url{https://api.semanticscholar.org/CorpusID:269214525}.

\bibitem[Tunstall et~al.(2023)Tunstall, Beeching, Lambert, Rajani, Rasul, Belkada, Huang, von Werra, Fourrier, Habib, Sarrazin, Sanseviero, Rush, and Wolf]{tunstall2023zephyr}
Lewis Tunstall, Edward Beeching, Nathan Lambert, Nazneen Rajani, Kashif Rasul, Younes Belkada, Shengyi Huang, Leandro von Werra, Clémentine Fourrier, Nathan Habib, Nathan Sarrazin, Omar Sanseviero, Alexander~M. Rush, and Thomas Wolf.
\newblock Zephyr: Direct distillation of lm alignment, 2023.

\bibitem[Wang et~al.(2024)Wang, Kulikov, Golovneva, Yu, Yuan, Dwivedi-Yu, Pang, Fazel-Zarandi, Weston, and Li]{Wang2024SelfTaughtE}
Tianlu Wang, Ilia Kulikov, Olga Golovneva, Ping Yu, Weizhe Yuan, Jane Dwivedi-Yu, Richard~Yuanzhe Pang, Maryam Fazel-Zarandi, Jason Weston, and Xian Li.
\newblock Self-taught evaluators.
\newblock 2024.
\newblock URL \url{https://api.semanticscholar.org/CorpusID:271709606}.

\bibitem[Wang et~al.(2022)Wang, Kordi, Mishra, Liu, Smith, Khashabi, and Hajishirzi]{Wang2022SelfInstructAL}
Yizhong Wang, Yeganeh Kordi, Swaroop Mishra, Alisa Liu, Noah~A. Smith, Daniel Khashabi, and Hannaneh Hajishirzi.
\newblock Self-instruct: Aligning language models with self-generated instructions.
\newblock In \emph{Annual Meeting of the Association for Computational Linguistics}, 2022.
\newblock URL \url{https://api.semanticscholar.org/CorpusID:254877310}.

\bibitem[Wu et~al.(2024{\natexlab{a}})Wu, Yuan, Golovneva, Xu, Tian, Jiao, Weston, and Sukhbaatar]{Wu2024MetaRewardingLM}
Tianhao Wu, Weizhe Yuan, Olga Golovneva, Jing Xu, Yuandong Tian, Jiantao Jiao, Jason Weston, and Sainbayar Sukhbaatar.
\newblock Meta-rewarding language models: Self-improving alignment with llm-as-a-meta-judge.
\newblock 2024{\natexlab{a}}.
\newblock URL \url{https://api.semanticscholar.org/CorpusID:271533411}.

\bibitem[Wu et~al.(2024{\natexlab{b}})Wu, Sun, Yuan, Ji, Yang, and Gu]{sppo}
Yue Wu, Zhiqing Sun, Huizhuo Yuan, Kaixuan Ji, Yiming Yang, and Quanquan Gu.
\newblock Self-play preference optimization for language model alignment.
\newblock 2024{\natexlab{b}}.

\bibitem[Xu et~al.(2023{\natexlab{a}})Xu, Sun, Zheng, Geng, Zhao, Feng, Tao, and Jiang]{xu2023wizardlm}
Can Xu, Qingfeng Sun, Kai Zheng, Xiubo Geng, Pu~Zhao, Jiazhan Feng, Chongyang Tao, and Daxin Jiang.
\newblock Wizardlm: Empowering large language models to follow complex instructions.
\newblock \emph{arXiv preprint arXiv:2304.12244}, 2023{\natexlab{a}}.

\bibitem[Xu et~al.(2023{\natexlab{b}})Xu, Guo, Duan, and McAuley]{xu2023baize}
Canwen Xu, Daya Guo, Nan Duan, and Julian McAuley.
\newblock Baize: An open-source chat model with parameter-efficient tuning on self-chat data.
\newblock \emph{arXiv preprint arXiv:2304.01196}, 2023{\natexlab{b}}.

\bibitem[Yin et~al.(2024)Yin, Wang, Xie, Chen, and Zhou]{sapo}
Yueqin Yin, Zhendong Wang, Yujia Xie, Weizhu Chen, and Mingyuan Zhou.
\newblock Self-augmented preference optimization: Off-policy paradigms for language model alignment.
\newblock \emph{ArXiv}, abs/2405.20830, 2024.
\newblock URL \url{https://api.semanticscholar.org/CorpusID:270199610}.

\bibitem[Yuan et~al.(2024)Yuan, Pang, Cho, Sukhbaatar, Xu, and Weston]{Yuan2024SelfRewardingLM}
Weizhe Yuan, Richard~Yuanzhe Pang, Kyunghyun Cho, Sainbayar Sukhbaatar, Jing Xu, and Jason Weston.
\newblock Self-rewarding language models.
\newblock \emph{ArXiv}, abs/2401.10020, 2024.
\newblock URL \url{https://api.semanticscholar.org/CorpusID:267035293}.

\bibitem[Zellers et~al.(2019)Zellers, Holtzman, Bisk, Farhadi, and Choi]{zellers2019hellaswag}
Rowan Zellers, Ari Holtzman, Yonatan Bisk, Ali Farhadi, and Yejin Choi.
\newblock Hellaswag: Can a machine really finish your sentence?
\newblock \emph{arXiv preprint arXiv:1905.07830}, 2019.

\bibitem[Zhang et~al.(2024)Zhang, Zhoubian, Yue, Dong, and Tang]{Zhang2024ReSTMCTSLS}
Dan Zhang, Sining Zhoubian, Yisong Yue, Yuxiao Dong, and Jie Tang.
\newblock Rest-mcts*: Llm self-training via process reward guided tree search.
\newblock \emph{ArXiv}, abs/2406.03816, 2024.
\newblock URL \url{https://api.semanticscholar.org/CorpusID:270285630}.

\bibitem[Zheng et~al.(2024)Zheng, Chiang, Sheng, Zhuang, Wu, Zhuang, Lin, Li, Li, Xing, et~al.]{mtbench}
Lianmin Zheng, Wei-Lin Chiang, Ying Sheng, Siyuan Zhuang, Zhanghao Wu, Yonghao Zhuang, Zi~Lin, Zhuohan Li, Dacheng Li, Eric Xing, et~al.
\newblock Judging llm-as-a-judge with mt-bench and chatbot arena.
\newblock \emph{Advances in Neural Information Processing Systems}, 36, 2024.

\bibitem[Zhong et~al.(2022)Zhong, Snell, Klein, and Steinhardt]{zhong2022describing}
Ruiqi Zhong, Charlie Snell, Dan Klein, and Jacob Steinhardt.
\newblock Describing differences between text distributions with natural language.
\newblock In \emph{International Conference on Machine Learning}, pp.\  27099--27116. PMLR, 2022.

\end{thebibliography}
\bibliographystyle{iclr2025_conference}

\clearpage

\section*{Limitations}
Our approach begins with a small SFT dataset and does not necessitate specifically labeled data for training a response improver. We assume that model-generated outputs closely resemble the gold standard responses in the SFT, enabling them to serve as training data for the response improver. This necessitates filtering out pairs where the gold standard is inferior to the model-generated output. Such data can cause the response improver to rewrite text through paraphrasing or substantial alteration, as the training data comprises pseudo pairs rather than minimally edited original responses. While prompting a more powerful LLM to generate rewriting-specific data, as suggested by \citet{jqself}, can alleviate this, it sacrifices the benefit of learning the distribution gap. 

SynPO leverages a small, high-quality dataset repeatedly to guide synthetic data generation, making the seed data quality vital. This approach requires only a small amount of high-quality data for validation, significantly reducing annotation costs. Additionally, recent work on direct on-policy sampling methods \citep{sppo}, which do not need additional SFT data, shows considerable promise. After our final round of improvements, the model-generated responses are already of high quality. Future enhancements can incorporate on-policy preference optimization techniques to further refine the model.

\appendix
\section*{Appendix}

\section{Algorithm}
\label{app:synpo}
We provide the overall pipeline of SynPO in Algorithm~\ref{alg:synpo}.
\begin{algorithm}[H]
\setcounter{ALC@unique}{0}
\caption{\texttt{\algname (SynPO)}}
\label{alg:synpo}
\resizebox{\textwidth}{!}{%
\begin{minipage}{\textwidth} 
\begin{algorithmic}[1]  
    \STATE \textbf{Input}: Initial policy $\pi_{\btheta_0}$, validation set $\left\{(\mathbf{x}^*_i, \mathbf{y}^*_i)\right\}_{i=0}^n$, keyword list set $\mathcal{K}$, prompt generator $\mathcal{G}$, data filter $\mathcal{F}$, synthetic preference data $\mathcal{D} = \emptyset$  
    \FOR{$t=1,2,\dots$}  
        \STATE Generate $m$ synthetic prompts $\{\mathbf{x}_i\}_{i=1}^{m}$ with $\mathbf{x}_i \sim \mathcal{G}(\cdot|\mathbf{k}_i)$, where $\mathbf{k}_i$ represents a list of keywords randomly sampled from $\mathcal{K}$. \label{line:generate}  
        \STATE Train the response improver $\mathcal{R}_t$ from $\btheta_0$, $\mathcal{R}_t \leftarrow \argmin_{\btheta} \sum_{i=0}^n \mathcal{L}(\pi_{\btheta}(\mathbf{x}^*_i, \mathbf{y}^*_{(t-1),i}), \mathbf{y}^*_i)$, where $\mathbf{y}^*_{(t-1),i} \sim \pi_{\btheta_{t-1}}(\cdot|\mathbf{x}^*_i)$ for $i \in \{1, \dots, n\}$.  
        \STATE Generate $\pi_{\btheta_{t-1}}$ completions and self-refined completions on the synthetic prompts: $\mathbf{y}_{(t-1),i} \sim \pi_{\btheta_{t-1}}(\cdot|\mathbf{x}_i)$, $\overline{\mathbf{y}_{(t-1),i}} \sim \mathcal{R}_t(\cdot|\mathbf{x}_i, \mathbf{y}_{(t-1),i})$ for $i \in \{1, \dots, m\}$.  
        \STATE Filter out invalid refinements and integrate valid data into the synthetic preference dataset:  
        \begin{align*}  
            \mathcal{D} \leftarrow \mathcal{D} \cup \left\{ (\mathbf{x}_i, \overline{\mathbf{y}_{(t-1),i}}, \mathbf{y}_{(0),i}) \mid \mathcal{F}(\mathbf{x}_i, \overline{\mathbf{y}_{(t-1),i}}, \mathbf{y}_{(0),i}) = \text{valid}, \ i \in \{1, \dots, m\} \right\}, \mathbf{y}_{(0),i} \sim \pi_{\btheta_0}(\cdot|\mathbf{x}_i)
        \end{align*}  
        \STATE Optimize $\pi_{\btheta_{0}}$ using the SimPO~\citep{meng2024simpo} objective, where $\sigma$ and $\gamma$ are hyperparameters: \label{line:optimize}  
        \begin{align*}  
            \btheta_{t} \leftarrow \argmin_{\btheta} \mathbb{E}_{(\mathbf{x}_i, \overline{\mathbf{y}_{(t-1),i}}, \mathbf{y}_{(0),i}) \sim \mathcal{D}} \left[  
                \log \sigma \left(  
                    \frac{\beta}{|\overline{\mathbf{y}_{(t-1),i}}|} \log \pi_\theta (\overline{\mathbf{y}_{(t-1),i}} \mid \mathbf{x}_i)  
                    - \frac{\beta}{|\mathbf{y}_{(0),i}|} \log \pi_\theta (\mathbf{y}_{(0),i} \mid \mathbf{x}_i)  
                    - \gamma  
                \right)  
            \right]  
        \end{align*}  
    \ENDFOR  
\end{algorithmic} 
\end{minipage}
}
\end{algorithm}

\section{Prompt for Response-refiner}
\label{app:rwt_prompt}

The prompt template used for training and inference in the response improver is shown in Figure \ref{prompt:rwt_prompt}.

\begin{figure}[h]
    \centering
    \input{rwt_prompt}
    \caption{Prompt in SynPO for the LLM to act as a response-refiner.}\label{prompt:rwt_prompt}
\end{figure}

\section{Experimental Details}
Here we list additional experimental details for our implementation and experiments.
\label{app:exp}
\subsection{Self-Prompt Generator Training}
The hyperparameters for self-prompt generator training are detailed below. During SFT for the self-prompt generator, we employ a learning rate of $1.0 \times 10^{-6}$ for Mistral-Base and Llama3-Base, with a batch size of 32, a warm-up ratio of 0.1, and an AdamW optimizer. We set the maximum sequence length to 8,000 and train the model for 3 epochs.   
  
To generate diverse synthetic prompts, we randomly sampled 1 million paragraphs from RefinedWeb~\citep{refinedweb} and randomly selected 3 keywords from each paragraph. This process yields a large keyword list pool containing 1 million keyword lists. These keyword lists serve as the input for the self-prompt generator in each iteration. For each iteration, we generate between 36,000 and 72,000 keyword lists (depending on the filtering ratio at each iteration) and exclude lists containing personal names or stopwords. We use vllm for inference and set the sampling temperature to 0.7.

\subsection{Response Improver Training}
As the model iterates and self-improves, it may produce responses superior to those of the seed data. Our objective is for the response improver to learn from its deficiencies. Therefore, we identify instances where the model output is inferior to the original response using the same scoring model as the filtering stage. This ensures that the response improver only learns positive optimizations or semantic paraphrasing, rather than negative optimizations.
Specifically, for a given $\mathcal{x}_i$, if the score difference between the gold standard completion and the model completion exceeds the threshold, we include this data for response improver training.  
In the Mistral-Base setting, we set the PairRM scoring threshold to 0.20. In the Llama3-Base setting, the ArmoRM-Llama3-8B-v0.1 scoring threshold is set to 0.02. 
  
Since the response improver data are automatically derived from SFT data conversion, the model also learns paraphrasing. Using a more powerful model, such as GPT-4, to create data that introduce only minor improvements for rewriter training is a promising research direction. However, to explore the potential for self-boosting, we did not introduce additional data or stronger models for data construction, resulting in inevitable paraphrasing by the response improver.

During SFT for the response improver, most training parameters are same to the parameters in self-prompt generator training. Some miner differences lie in: we set the max sequence length to 6,000.

\subsection{Response Improving and Filtering Setting}

To produce synthetic preference (chosen and rejected) completions for the $t$-th iteration, we utilize the current policy model to generate a completion and employ the response improver to refine it. We use vllm for inference, with the decoding temperature set at $T = 0.7$. 

During synthetic data filtering, we set a threshold 0.20 for PairRM scores and a threshold 0.02 for ArmoRM-Llama3-8B-v0.1 scores. 
In addition, we filter out all the data that contain over 50\% repetition patterns to avoid model collapse on synthetic data.
In our experiments, we randomly incorporated 10,000 preference pairs from each iteration to the whole synthetic preference data.

\subsection{Optimization}
As the parameter $\beta$ is crucial for achieving optimal performance in SimPO~\citep{meng2024simpo}, we individually search the $\beta$ in the range of [2, 4, 6, 8, 10, 12] for each optimization process. We use a fixed $\gamma = 1.6$ for the Mistral-Base model and Llama3-Base.

\subsection{Baselines}
\label{app:bsln}
In experiments involving iterative baselines, we control various conditions to ensure fairness. We maintain the same training data size for both iterative baselines and SynPO. We adopt the SimPO loss~\citep{meng2024simpo} for preference optimization, as it is more effective than DPO~\citep{dpo}. 
We all use self-generated prompts, which have been shown to be superior to prompts generated by other methods, as validated in Section~\ref{sec:similarity-distribution} and Section~\ref{sec:prompt-ablt}.
All preference construction processes are iterated until performance no longer improves.  

Regarding the baseline models trained on UltraFeedback 61k, we straightforwardly adopt the well-trained versions available from the SimPO repository at \url{https://github.com/princeton-nlp/SimPO}.

\subsection{Decoding Hyperparameters}
For the AlpacaEval 2~\citep{alpacaeval} evaluation, we use a sampling-based decoding approach to generate responses. Specifically, we employ vllm for inference, setting the temperature to 0.7 and the maximum tokens to 2048 for both the Mistral-Base and Llama3-Base configurations. All other parameters adhere to the default settings in vllm.
As for MT-Bench~\citep{mtbench}, we adhere to the official decoding setup, which specifies varying sampling temperatures tailored to distinct categories.

\section{Additional Details on Seed Data Ablation}
\label{app:seed}
In setting 2), 3), and 4), to prevent cases where rejected responses are better than the chosen ones, we filter the preference data using the same method as SynPO, specifically employing ArmoRM-Llama3-8B-v0.1 to select valid preference data. We fix the threshold at 0.02, as our search among \{0, 0.1, 0.2\} reveal that 0.02 consistently performs the best.

\section{API Usage}
\label{app:api}
For GPT-4 Turbo, we all use the latest turbo-2024-04-09 API on Azure OpenAI Service~\url{https://learn.microsoft.com/en-us/azure/ai-services/openai/concepts/models#gpt-4-turbo}.

\section{Prompt Analysis}
\label{app:prompt-analysis}

\begin{figure}[ht]
    \includegraphics[width=1\linewidth]{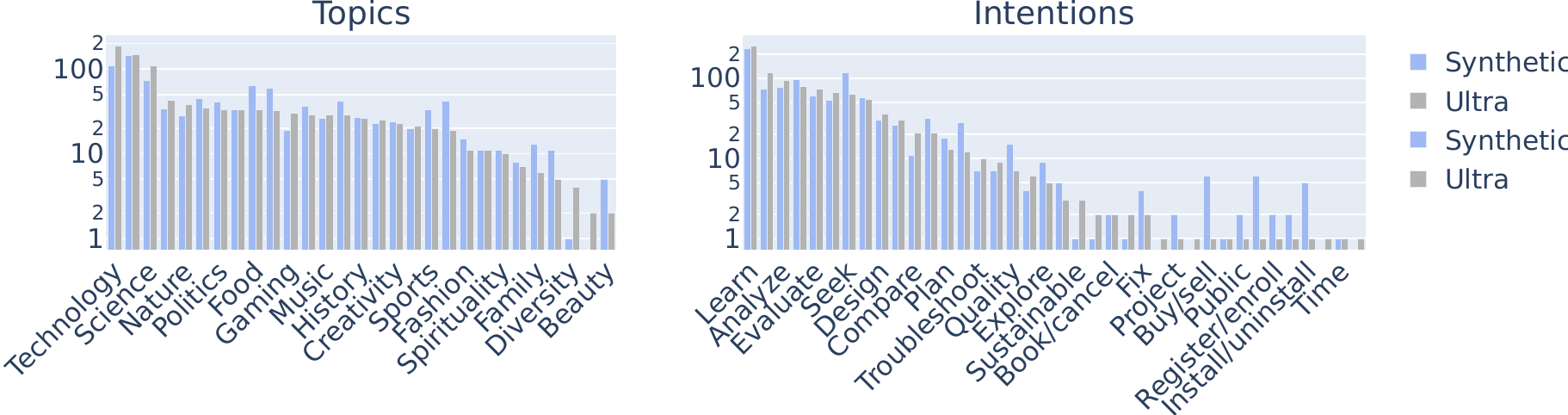}
    \caption{Topic and intention distribution bar plot for the synthetic prompts in SynPO and UltraFeedback.}
    \label{fig:questionanalysis-bar}
\end{figure}

Here we provide the prompt used for prompt topic and intention analysis in Figure~\ref{fig:cls_prompt}, along with a more detailed distribution bar plot for different intentions and topics in Figure~\ref{fig:questionanalysis-bar}. The topic word list is derived from UltraChat~\citep{ding2023enhancing}, while the intention word list was designed by us.  

\begin{figure}[ht]
    \centering
    \input{cls_prompt}
    \caption{Prompt for using GPT-4 Turbo as a intention and topic classifier.}
    \label{fig:cls_prompt}
\end{figure}

\section{Evaluation Details}
\label{app:eval}

\begin{figure}[t]
    \centering
    \includegraphics[width=1\linewidth]{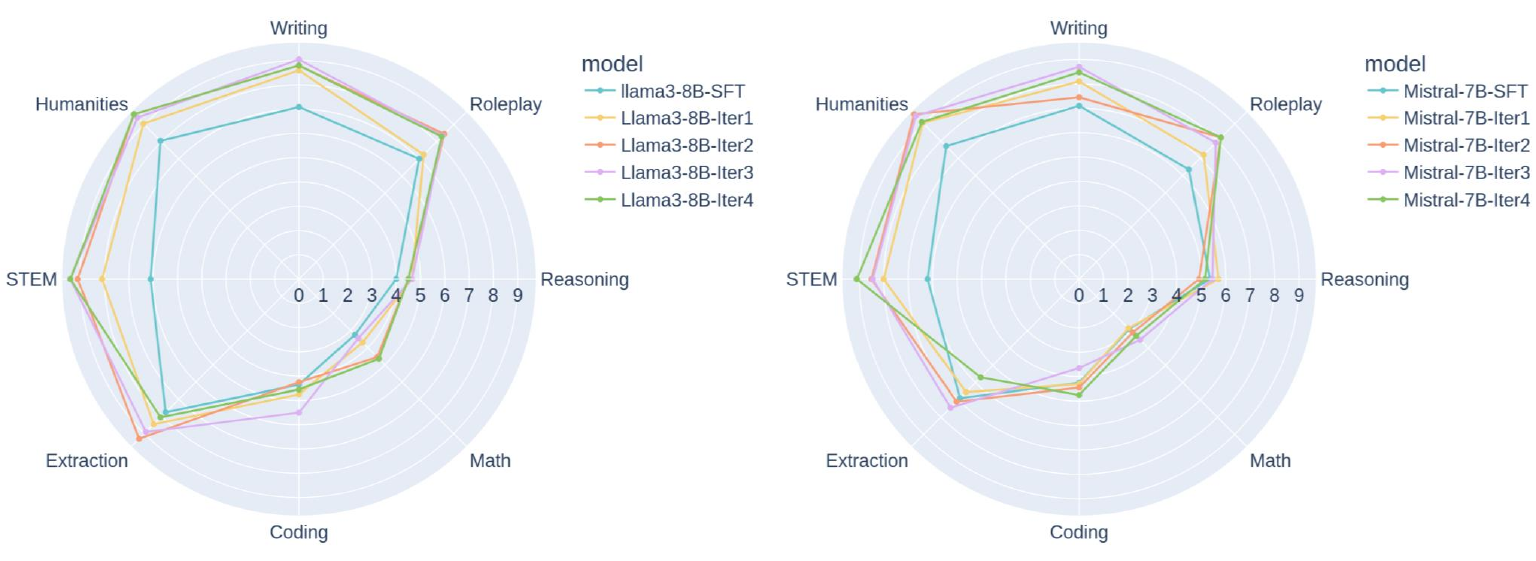}
    \caption{MT-Bench scores on different prompt types. The left radar chart represents results from SynPO on Llama3 and the right comes from Mistral.}\label{app:mtbenchradar}
\end{figure}

For instruction-following ability evaluation, 
Table~\ref{tab:alignbench} presents the detailed information for three alignment benchmarks we use, including AlpacaEval 2.0, Arena-Hard and MT-Bench. Additionally, we display the radar chart for MT-Bench scores on different prompt types (see Figure~\ref{app:mtbenchradar}).

As for general LLM capability evaluation, we provide the few-shot example numbers on Open LLM Leaderboard in
Table~\ref{tab:num-fewshot} 
and a comprehensive comparison of SynPO.

\setlength{\tabcolsep}{3pt}
\begin{table*}[h]
\label{tab:eval_dataset}
\centering
\resizebox{0.9\textwidth}{!}{
\begin{tabular}{@{}lcccc@{}}
\toprule
& \textbf{\# Instances} & \textbf{Baseline Model} & \textbf{Judge Model}           & \textbf{Scoring Type}  \\ \midrule
AlpacaEval 2.0 & 805                   & GPT-4 Turbo      & GPT-4 Turbo       & Pairwise comparison    \\
Arena-Hard    & 500                   & GPT-4-0314              & GPT-4 Turbo       & Pairwise comparison   \\  
MT-Bench     & 80                    & -                       & GPT-4 Turbo & Single-answer grading  \\
\bottomrule
\end{tabular}
}
\caption{Details for three alignment benchmarks. 
}
\label{tab:alignbench}
\vspace{-1.2em}
\end{table*}
\setlength{\tabcolsep}{6pt}

\setlength{\tabcolsep}{4pt}
\begin{table}[ht]
    \centering
     \begin{tabular}{l|rrrrrr}
        \toprule                
       \bf  Task   & \bf  Arc & \bf HellaSwag & \bf  TruthfulQA & \bf MMLU & \bf Winogrande & \bf GSM8k    \\
        \midrule
    \# Few-shot Examples  & 25 & 10 & 0 & 5 & 5 & 5\\
    Metrics & acc\_norm & acc\_norm & mc2 & acc & acc & acc\\
        \bottomrule
        \end{tabular} 
    \caption{Number of few-shot examples in Open LLM Leaderboard evaluation.}\label{tab:num-fewshot}
    \label{}
\end{table}

\end{document}